\documentclass{article}
\usepackage{ijcai26}

\usepackage{times}
\usepackage{soul}
\usepackage{url}
\usepackage[hidelinks]{hyperref}
\usepackage[utf8]{inputenc}
\usepackage{graphicx}
\usepackage{amsmath}
\usepackage{amsfonts}
\usepackage{amsthm}
\usepackage{booktabs}
\usepackage{algorithm}
\usepackage{algorithmic}
\usepackage[switch]{lineno}
\usepackage{subcaption} 
\usepackage{subcaption}
\usepackage{float} 
\usepackage{multirow}
\usepackage{hyperref}
\usepackage{placeins}

\title{Climate Surrogates for Scalable Multi-Agent Reinforcement Learning: A Case Study with CICERO-SCM}

\author{
Oskar Bohn Lassen$^1$
\and
Serio Agriesti$^1$\and
Filipe Rodrigues$^1$\and
Blaz Kurnik$^2$ \and
Francisco Pereira$^1$\\
\affiliations
$^1$Technical University of Denmark\\
$^2$European Environment Agency\\
\emails
\{obola, samaa, rodr\}@dtu.dk,
blaz.kurnik@eea.europa.eu,
camara@dtu.dk
}

\begin{document}

\maketitle

\begin{abstract}
Climate policy analysis requires models that capture multi-gas climate effects, but such models are too slow to embed in reinforcement learning loops at scale. 
In collaboration with the European Environment Agency, we develop a multi-agent reinforcement learning (MARL) framework that integrates a higher-fidelity climate surrogate as the environment transition, enabling regional agents to learn policies under multi-gas dynamics. 
We train a recurrent surrogate on $20{,}000$ multi-gas emission pathways to emulate CICERO-SCM. 
The surrogate achieves near-simulator accuracy (global-mean temperature RMSE $\approx\!4\!\times \!10^{-4}\,\mathrm{K}$) with $\sim\!1000\times$ faster one-step inference and yields $>\!100\times$ end-to-end MARL training speed-up. 
We show policy agreement with the simulator in tractable settings and propose a replay- and rank-consistency test (Kendall's $\tau$) for assessing policy fidelity when simulator-in-the-loop training is infeasible.
This enables large-scale multi-agent policy experiments while retaining high-fidelity multi-gas climate response.
\end{abstract}

\section{Introduction} \label{sec:introduction}
Climate modeling provides the backbone for climate policy exploration, but there is an inherent trade-off between model fidelity and computational cost. 
State-of-the-art Earth System Models (ESMs) simulate climate processes in great detail \cite{esm2018overview}, but a single ESM run can take days to weeks on high-performance computers. 
This extreme cost severely limits how many scenarios or policy strategies can be evaluated, motivating the use of simpler models for many applications~\cite{scm2025iam}.

The climate science community relies on reduced-complexity Simple Climate Models (SCMs) to emulate climate responses at far lower computational cost. SCMs are compact models (often energy-balance models with simplified ocean and carbon-cycle components) calibrated to reproduce the behavior of more complex ESMs~\cite{scm2020overview}. 
For example, widely used SCMs like MAGICC \cite{meinshausen2011magicc} and CICERO-SCM \cite{fuglestvedt1999cicero} provide tractable climate projections using simplified physics, while the FaIR model takes an impulse-response approach~\cite{smith2018fair,scm_comparison}. 
These models sacrifice some process-level detail for high computational efficiency. 
Because SCMs run orders of magnitude faster than ESMs, they are widely used in applications that require large ensembles or iterative evaluations, such as probabilistic climate projections or integrated assessment models (IAMs)~\cite{scm2025iam}.

IAMs couple socio-economic systems with a climate module to assess mitigation and impact pathways, translating climate outcomes into economic metrics. 
Pioneering IAMs like DICE and RICE used highly simplified climate modules in an optimal growth framework~\cite{nordhaus1992dice,nordhaus1993dice,nordhaus2014dice2013r,nordhaus1996rice,nordhaus2010rice,nordhaus2016rice}. 
Building on this foundation, more detailed IAM frameworks (e.g., REMIND-MAgPIE, MESSAGE-GLOBIOM, WITCH, IMAGE) introduced additional realism (e.g. technological detail, land-use, energy systems) \cite{kriegler2017remind,fricko2017message,emmerling2016witch,stehfest2014image}, often exploring socio-economic scenario families (SSPs)~\cite{riahi2017ssp}. 
In all cases, IAMs rely on fast climate simulators like MAGICC \cite{meinshausen2011magicc} because ESMs are too slow for large scenario ensembles or iterative policy optimization loops. 
However, traditional IAMs typically model the world as a few aggregate regions optimizing an equilibrium trajectory under strong foresight. 
This aggregation and reliance on optimal-control or game-theoretic assumptions can mask heterogeneity in preferences, adaptive behaviors, and path-dependent collective dynamics. 
These limitations have prompted interest in more flexible, simulation-based approaches~\cite{iamsnotenough,agentbased2023iam}.

Reinforcement learning (RL), especially multi-agent RL (MARL), has been proposed as an alternative framework for studying climate-economy interactions~\cite{strnad2019deep,ruddjones2025multi}. 
In an MARL formulation, multiple agents (e.g. countries) make simultaneous decisions and learn strategies through repeated interaction in a simulated environment, rather than assuming a globally optimized equilibrium path. 
This approach accommodates heterogeneous agents, non-linear dynamics, and bounded rationality. 
An early attempt to apply MARL in an IAM context is RICE-N \cite{zhang2022ricen}, which replaced RICE's decision-making with learning agents. 
However, RICE-N used an overly simplified climate module and limited cooperation mechanisms, restricting its realism and the policies that could emerge. 
Subsequent works began to address these gaps:\cite{ruddjones2025crafting} incorporated more sophisticated cooperation and coalition formation, and \cite{heitzig2023commitments} examined the impact of commitment strategies on climate outcomes. 
The recent JUSTICE framework \cite{biswas2025justice} went further by integrating the FaIR climate model into a multi-objective MARL game, improving the fidelity of climate dynamics in the learning environment. 
Nonetheless, even JUSTICE only allowed control of CO$_{2}$ with a low-dimensional action space and a small number of agents, limiting exploration of complex policies and undercutting the benefits of a detailed climate response.

In reality, different greenhouse gases and aerosols have widely varying lifetimes and warming effects. 
Long-lived gases like CO$_2$ and N$_2$O persist for centuries (driving long-term warming), methane lasts around a decade (strong near-term warming), and short-lived pollutants (e.g. aerosols) persist for months and often have a cooling effect. 
Mitigation actions also influence these species heterogeneously: for example, decarbonizing the energy sector reduces CO$_2$ emissions but also lowers SO$_2$ aerosol emissions (removing some cooling and causing a short-term warming), while halting deforestation cuts CO$_2$ (land-use) and CH$_4$ without a similar short-term penalty. Thus, MARL climate games need a climate engine that responds to multi-gas emission patterns rather than just aggregate CO$_2$. 
To date, however, MARL studies have relied on oversimplified climate dynamics and severely restricted action spaces, mainly due to computational constraints.

Scalability is a critical issue: MARL experiments often require $10^7$ environment interactions for agents to learn effective policies~\cite{papoudakis2021benchmarking}. 
Even an SCM taking $\sim\,0.4$ seconds per step can become a bottleneck when millions of calls are needed. 
This is a key reason why prior MARL studies have not used complex or multi-gas climate models - doing so would make training intractable. 
One promising approach to embed higher-fidelity climate dynamics into MARL frameworks without incurring prohibitive cost is the use surrogate models.
Machine-learning based surrogates have accelerated expensive components of ESMs by orders of magnitude \cite{lu2019efficient}, and deep neural networks can mimic short-term climate projections with high speed and reasonable accuracy \cite{weber2020deep}. 
However, most prior climate surrogates target offline surrogate modeling, whereas in this setting, a surrogate must become a component of a MARL environment transition function, where small dynamics errors can alter exploration, coordination, and even the learned equilibrium.

In this paper, we extend the realism of MARL climate-policy games by embedding a more complex, multi-gas climate model into the environment while keeping training tractable. 
First, we propose a modular framework for substituting a surrogate model in place of the climate module within a climate-economy MARL environment. 
Second, we introduce a recurrent neural network surrogate for CICERO-SCM and describe our MARL experimental setup, including a method to evaluate policy consistency when running the simulator is infeasible. 
Finally, we present results demonstrating that our surrogate dramatically accelerates MARL training ($>\!100\times$ speed-up) without sacrificing policy fidelity.

The work has been developed in collaboration with the European Environment Agency where a domain expert have engaged with us to align the problem setting with policy-relevant use cases and operational constraints (e.g., realism of emission pathways and controllable levers) and have provided feedback on the experimental protocol and interpretation of results.

\begin{figure*}[t]
  \centering
  \includegraphics[width=0.9\linewidth,
                trim={0.9cm 1.0cm 0.8cm 0.8cm},
                clip]{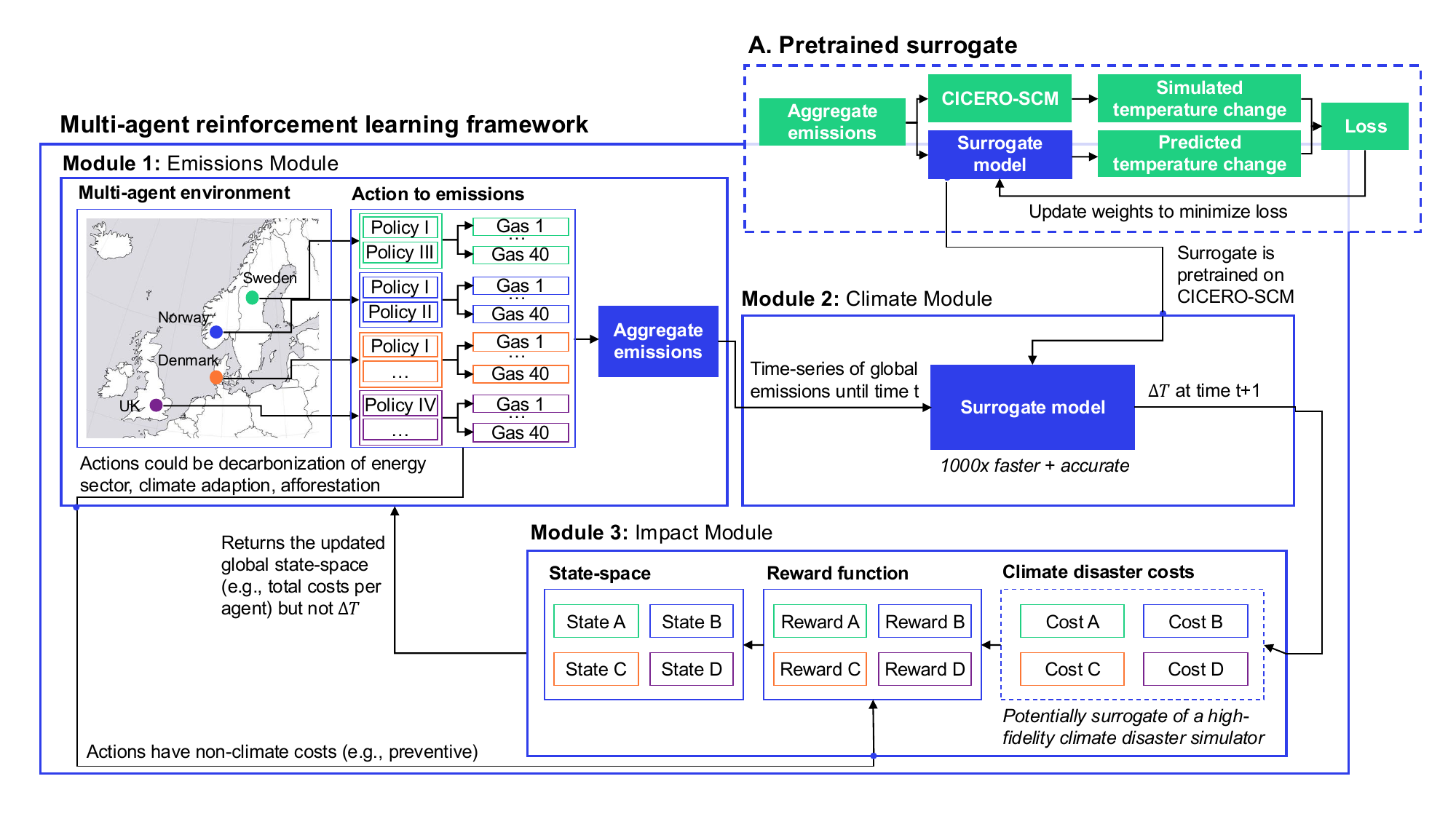}
  \caption{Framework for integrating climate surrogates into MARL. In Module 1, agents choose policies that result in emissions, which in Module 2 are mapped into temperature change by a pretrained surrogate and in Module 3 converted into costs.}
  \label{fig:marl_diagram}
\end{figure*}

\section{Conceptual Framework} \label{sec:framework}
According to the IPCC Sixth Assessment Report \cite{ipcc2022wg2spm,nabuurs2022afolu,clarke2022energy}, key intervention areas in climate policy include: (i) decarbonizing the energy sector (e.g.\ coal phase-out, renewable expansion), (ii) targeting methane abatement (e.g.\ waste management, leak mitigation), (iii) improving agricultural and land-use practices (e.g.\ reduced deforestation, fertilizer efficiency), and (iv) investing in adaptation or preventive measures that reduce realized damages. 
These policy levers affect multiple greenhouse gases, each with different climate effects. 
Ideally, the resulting emission changes should feed into a high-fidelity climate model to provide a realistic estimate of temperature increase, which then propagates into region-specific damage estimates.

However, there is a trade-off between environment complexity and tractability: simplifying the climate model makes experiments feasible but limits the policies and dynamics that can be explored. 
We propose a modular framework in which complex climate components can be replaced by fast surrogates. 
We divide the environment conceptually into three modules:

\begin{enumerate}
    \item \textbf{Emissions Module:} maps agent actions to multi-gas emissions
    \item \textbf{Climate Module:} maps emissions to climate response, e.g., temperature change
    \item \textbf{Impact Module:} maps climate outcomes and actions to costs or damages
\end{enumerate}

As illustrated in Figure~\ref{fig:marl_diagram}, one can seamlessly substitute a higher-fidelity climate model $f_{\mathrm{SCM}}$ with a learned surrogate model $f_{\theta}$, without altering any other part of the environment. 
The surrogate emulates the original model's behavior, preserving the multi-gas climate response fidelity while enabling the use of a high-speed, hardware-optimized model in the RL training loop.
This modular design also provides flexibility to increase fidelity in other parts of the environment without changing the learning algorithm. 
For instance, the impact module could be a simple global damage function or a detailed model of local impacts (e.g., regional sea-level rise or crop yield changes) without requiring any modifications to the agents’ decision-making loop.

In the remainder of this paper, we instantiate this framework using CICERO-SCM as our high-fidelity climate engine and a recurrent neural network (RNN) as the surrogate emulator. 
We then integrate the surrogate into a multi-agent climate-economic environment and demonstrate the resulting speed-ups and policy consistency.

\section{Methodology}
\label{sec:methodology}
In this section, the methodological steps adopted to design a surrogate for CICERO-SCM and embed it into the MARL framework are described.

\subsection{Climate Dynamics Engine: CICERO-SCM}
\label{sec:cicero}
CICERO-SCM (v1.1.1) \cite{sandstad2024cicero} is a reduced-complexity global climate model that maps multi-gas emission trajectories to global-mean surface air temperature. 
Let $G$ be the set of gases ($(|\mathcal{G}|=40)$) used in the model, and define the global emissions vector $E(t) = [E_g(t)]_{g\in\mathcal{G}} \in \mathbb{R}^{|\mathcal{G}|}$ and the emissions history $E_{1{:}t} = [E(\tau)]_{\tau=1}^{t} \in \mathbb{R}^{t\times|\mathcal{G}|}$.

At each year $t$, the simulator produces a temperature change $\Delta T(t) = f_{\mathrm{SCM}}\!\big(E_{1{:}t}\big)$, where $\Delta T(t)$ is the global mean surface air temperature change in year $t$.
Internally, CICERO-SCM couples a carbon-cycle module for CO$_2$, exponential-decay models for other long-lived gases (e.g., CH$_4$, N$_2$O,), and an upwelling-diffusion energy-balance model that translates total radiative forcing into temperature change \cite{etminan2016radiative}. 
Its parameters are calibrated to ESM outputs and observations, yielding a physically consistent climate response (Appendix~A.2). However, each call still takes $\sim\!0.4$ s, which would make millions of simulator steps (as required in MARL) intractable.

\subsection{Surrogate Model of CICERO-SCM}
\label{sec:surrogate}
\paragraph{Generation of Emission Trajectories for Model Training}
For surrogate training, we generated an ensemble of $S=20{,}000$ policy-relevant emission trajectories by randomly perturbing the year-over-year growth rates of a baseline scenario (SSP2-4.5 \cite{FRICKO2017251}) over the period 2015-2075. 
We apply smoothing to these perturbations to produce realistic trajectories that mimic the smoother patterns of learned policies. Specifically, we first compute the baseline annual growth factor for each gas $g$:
\begin{equation}
\label{eq:growth_base}
\delta_g^{\mathrm{base}}(t) = \frac{E_g^{\mathrm{base}}(t)}{E_g^{\mathrm{base}}(t-1)}
\end{equation}

For each scenario $s$, we draw multiplicative factors $\tilde{\zeta}^s_g(t) \sim \mathcal{U}(\ell_g, u_g)$ and then smooth them via an exponential moving average:
\begin{align}
\zeta_g^s(t) &= \bigl(\zeta_g^s(t{-}1)\bigr)^{\alpha}\,
                \bigl(\tilde{\zeta}_g^s(t)\bigr)^{1-\alpha},
& \zeta_g^s(2015) &= 1
\end{align}
These factors perturb the baseline growth, $\delta^s_g(t) = \delta^{\text{base}}_g(t) \,\zeta^s_g(t)$. 
This defines scenario-specific emissions recursively: for $t \le 2015$, $E^s_g(t) = E^{\text{base}}_g(t)$ (using historical emissions); for $t \ge 2016$, $E^s_g(t) = E^s_g(t-1)\,\delta^s_g(t)$.

\begin{figure}[t]
  \centering
  \includegraphics[width=0.48\textwidth]{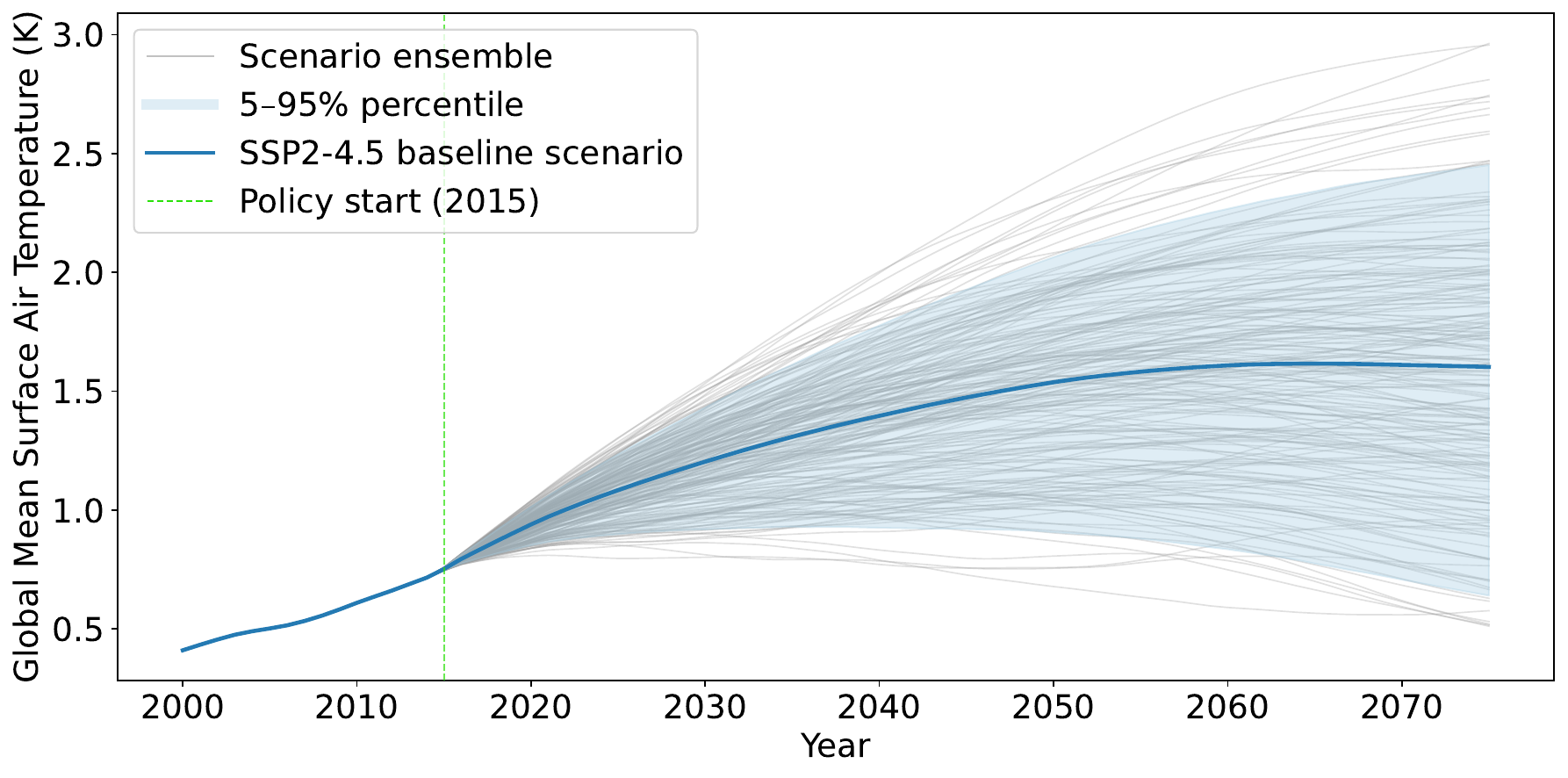}
  \caption{Global mean surface air temperature change for the generated emission trajectories.}
  \label{fig:temp_ensemble}
\end{figure}

We focus on a subset $\mathcal{C} \subset \mathcal{G}$ of five controllable gas species to be controlled in the MARL experiment (Section~\ref{sec:marl}): CO$_2$ from fossil fuels (FF), CO$_2$ from land-use (AFOLU), CH$_4$, N$_2$O, and SO$_2$. 
Perturbations are applied only to these gases by setting $(\ell_g, u_g) = (0.925,1.075)$ for $g \in C$ (allowing $\pm 7.5\%$ annual growth variation) and $(\ell_g, u_g) = (1, 1)$ for $g \notin C$ (those emissions follow the baseline). 
We choose $\pm 7.5\%$ to intentionally exceed the MARL agents’ maximum per-year control range ($\pm 5\%$; Appendix~D, Tables~A.3-A.4), so that all policy-induced emission trajectories in the intended MARL setting remain within the surrogate’s training support.
Thus, the surrogate is designed for accuracy on this feasible policy-induced emission space, rather than for unconstrained extrapolation far outside these bounds.
These perturbations are sampled independently across gases to ensure coverage of the controllable action subspace which avoids under-representing low-magnitude gases that could otherwise be overshadowed by correlated variations.
Running CICERO-SCM on all $20{,}000$ scenarios yields an ensemble of temperature trajectories (Figure~\ref{fig:temp_ensemble}). 
The spread of global temperature outcomes is narrow in the near term (since all scenarios share the same historical emissions) but grows substantially by 2075 as the scenarios diverge. 
The unperturbed SSP2-4.5 baseline lies near the median of the ensemble, indicating that our scenario generation spans futures consistent with that reference while covering substantial variability in warming outcomes.

For each scenario $s$ and year $t$, let $E^s(t) = [E^s_g(t)]_{g\in \mathcal{G}}$ be the full emission vector and $E^s_\mathcal{C}(t) = [E^s_g(t)]_{g\in \mathcal{C}}$ the vector of emissions for the controllable gases. We use the paired data $\{E^s_\mathcal{C}(t), \Delta T^s(t)\}$ to train the surrogate:
\begin{align}
X^s(t) = \bigl[E^s_{\mathcal{C}}(t{-}W),\,\dots,\,E^s_{\mathcal{C}}(t)\bigr] , \quad y^s(t) = \Delta T^s(t)
\end{align}
where $W$ is the length of the temporal window.
We choose $W = 65$ years so that the input covers sufficient historical context, to capture slow climate responses and long-lived gas effects, while remaining computationally manageable.
We generate one sample for each scenario and each target year $t \in \{2015, \dots, 2075\}$ (excluding years before 2015 since all scenarios share identical emissions). 
This yields 61 samples per scenario, for a total of about $1.22$ million training examples. 
The data are split by scenario into 70\% training, 15\% validation, and 15\% testing, ensuring that no scenario’s future data appear in its training set.

\paragraph{Architecture of RNN-based Surrogate Model}
We build a recurrent neural network surrogate comprising three modules: (i) an RNN-based encoder, (ii) a skip-connection that feeds current-year emissions directly to the predictor, and (iii) a two-layer feedforward prediction head. 
The RNN encoder processes the historical emissions window $[x_{t-W}, \dots, x_{t-1}]$, where $x_{\tau} \in \mathbb{R}^{|\mathcal{C}|}$ is the controllable-gas emissions at time $\tau$, and produces a hidden state $h_t$ summarizing the past dynamics. 
We experimented with different sequence models for this encoder, including Long Short-Term Memory (LSTM) \cite{hochreiter1997lstm}, Gated Recurrent Unit (GRU) \cite{gru2014}, and a Temporal Convolutional Network (TCN) \cite{lea2016temporalconvolutionalnetworksaction} (which replaces recurrence with convolutions while preserving the input-output structure). 
These architectures are well suited to the annual, low-dimensional emission sequences considered here, while keeping inference fast inside the MARL loop.
A skip-connection then concatenates the current-year emissions $x_t$ with the encoded history to form $z_t = [h_t; x_t]$, ensuring that recent emission changes are directly available to the predictor.
Finally, the prediction head - a two-layer multilayer perceptron - maps $z_t$ to the predicted temperature change $\Delta \hat T(t)$. 
The surrogate is not autoregressive in temperature: previous temperature predictions are not fed back as inputs, which limits classical forecast-drift effects.

\subsection{MARL Climate Mitigation Experiment}
\label{sec:marl}
We formulate a finite-horizon Markov game played annually from 2016 to 2050 ($H=35$ time steps) among $N$ agents (representing countries). 
Each year $t$, all agents choose actions simultaneously, and these actions determine the emissions and resulting climate outcomes for the next step.
The environment is intentionally stylized to isolate the surrogate-integration and policy-fidelity questions, rather than to serve as a complete IAM.

\paragraph{Actions}
Each agent $i$'s action is a tuple $a_{i,t} = (e_{i,t}, m_{i,t}, l_{i,t}, p_{i,t})$, where $e_{i,t} \in \{0, 0.5, 1.0\}$ is the effort in decarbonizing energy, $m_{i,t}$ is the effort in methane abatement, $l_{i,t}$ is the effort in land-use and agriculture measures, and $p_{i,t} \in \{0, 0.03, 0.08\}$ is the fraction of output invested in adaptation (preventive measures). 
The three mitigation levers $(e, m, l)$ affect the growth rate of emissions in $C$ via a fixed mapping matrix $M\in\mathbb{R}^{3\times|\mathcal{C}|}$ (each action influences each gas in $C$ by a predetermined amount).
Given agent i's mitigation effort vector $k_{i,t}=(e_{i,t},m_{i,t},l_{i,t})$ the induced growth-rate deviation for gas g is defined as:
\begin{align}
\tilde\delta_{i,g}(t) &=
\begin{cases}
[k_{i,t}\, M]_g & g\in\mathcal{C}\\[6pt]
0 & g\in \mathcal{G}\setminus \mathcal{C}
\end{cases}
\end{align}
and the effective growth factors are then given by:
\begin{align}
\delta^{\mathrm{eff}}_{i,g}(t) = \delta^{\mathrm{base}}_g(t)\,\big(1+\tilde\delta_{i,g}(t)\big) \label{eq:growth}
\end{align}
The adaptation action $p_{i,t}$ accumulates into a prevention stock $P_i$ that multiplicatively attenuates climate damages in the reward function.

Each agent's emissions are initialized in 2015 as $\bar{E}_i(2015) = S_i \odot E^{\text{base}}(2015)$, where $S_i$ is the vector of the agent’s share of baseline emissions (with $\sum_i S_i = \mathbf{1}$). 
Thereafter, given the agent’s effective emission growth factors (after applying its actions), the per-agent emissions are updated each year ($\bar{E}_i(t) = \bar{E}_i(t-1) \odot \delta^{\text{eff}}_{i}(t)$), so that the global emissions are $E(t) = \sum_{i=1}^N \bar{E}_i(t)$. 
The climate module (either the simulator $f_{\text{SCM}}$ or the surrogate $f_{\text{NET}}$) then takes the aggregated emissions and produces the global temperature change $\Delta T(t)$. 
The surrogate uses only the controllable subset $E_{\mathcal{C}}(t)$ as input, whereas the full simulator uses the entire $E(t)$; both are updated iteratively and carry state forward in time.

\paragraph{Observation}
Each agent receives the same global observation $O(t)$:
\begin{align}
\begin{aligned}
O(t)=\big[&
\Delta T(t-1),\tau(t),\bar E_i^{\mathcal{C}}(t{-}1),\\
& D_i^{\mathcal{C}}(t{-}1), P_i(t{-}1)
\big]
\end{aligned}
\end{align}
where $\tau(t)\in[0,1]$ is a normalized year index, $\bar E_i^{\mathcal C}(t{-}1)$ are last-year realized emissions for the controllable gases ($\mathcal C$) per agent, $D_i^{\mathcal C}(t{-}1)$ are cumulative deviations from baseline emissions for the controllable gases ($\mathcal C$) per agent, and $P_i(t{-}1)\in[0,P_{\max}]$ are prevention stocks per agent. 
The two summary quantities are defined as:
\begin{align}
D_i^{\mathcal C}(t)&=\sum_{u=2016}^{t}\Big(\bar E_i^{\mathcal C}(u)-S_i^{\mathcal C}\odot E^{\mathrm{base},\mathcal C}(u)\Big)\\
P_i(t)&=\min\!\big\{P_{\max},\, P_i(t{-}1)\, \phi\,+p_{i,t}\big\}
\end{align}
where $P_{\max}$ is the maximum effect prevention can have on the climate cost and $\phi \in [0,1]$ is a decay rate on the preventive investments. 

\paragraph{Reward}
At each year $t$, the reward for agent $i$ is the negative of three types of costs: $C_{i}^{\mathrm{c}}(t)$ climate disaster costs, $C^{\mathrm{k}}_{i}(t)$ policy costs, and $C^{\mathrm{p}}_{i}(t)$ prevention costs:
\begin{align}
r_i(t) &= -\,\eta\Big( C_i^{\mathrm{c}}(t) \;+\; C_i^{\mathrm{k}}(t) \;+\; C_i^{\mathrm{p}}(t) \Big) \label{eq:reward_start}\\
C_i^{\mathrm{c}}(t) &= c_i^{\mathrm{c}} \,\psi\,\big(\Delta T(t)\big)^{4}\,\bigl(1 - P_i(t)\bigr)\\
C_i^{\mathrm{k}}(t) &= \big(c_i^{\mathrm{k}}\big)^\top \!\big( k_{i,t} \odot k_{i,t} \big) \\
C_i^{\mathrm{p}}(t) &= c_i^{\mathrm{p}} \cdot p_{i,t} \label{eq:reward_end}
\end{align}
where $c_i^{\mathrm{c}}$, $c_i^{\mathrm{k}}=(c_i^{\mathrm{e}},c_i^{\mathrm{m}},c_i^{\mathrm{l}})$, and $c_i^{\mathrm{p}}$ are the agent-specific climate cost, policy costs, and prevention cost respectively. $P_i(t)\in[0,P_{\max}]$ is the prevention stock, $\psi = 0.003$ is the base climate damage calibration parameter, and $\eta=10^{-1}$ is the coefficient for reward normalization. 

At the terminal step we add a look-ahead cost for near-term climate damage. 
From the terminal state, $t\!=2050$, we roll the climate model forward for $U=15$ years with baseline emission growth $\delta_g^{\text{base}}(t)$ used to calculate $\bar E(t)$ and decaying prevention $P_i(t{+}u)=\min\{P_{\max},\,P_i(t)\,\phi^u\}$ for $[u,\dots,U]$:
\begin{align}
r_i^{\mathrm{term}}(t)
&=
\sum_{u=1}^{U} \ C_i^{\mathrm{c}}(t{+}u) \\
r_i(t) &\leftarrow r_i(t)\;-\; r_i^{\mathrm{term}}(t)\,\eta
\end{align}
This terminal look-ahead serves as a pragmatic reward-shaping term to reflect imminent damages in the terminal step, rather than as a general solution to long-horizon credit assignment.

\paragraph{Optimization}
We train independent recurrent policies using Proximal Policy Optimization (PPO) \cite{schulman2017ppo}.
Each agent $i$ has parameters $\theta_i$ and seeks to maximize its expected discounted return:
\begin{align}
\theta_i^{\star}
&= \arg\max_{\theta_i}\;
J_i(\theta_i) \\
J_i(\theta_i)
&= \mathbb{E}_{\tau \sim \pi_{\theta_i}}
   \!\left[\sum_{t=0}^{H-1}\gamma^{t}\,r_i(t)\right],
   \qquad \gamma = 0.999
\end{align}
where $r_i(t)$ is the per-step reward defined in eqs.~(\ref{eq:reward_start}–\ref{eq:reward_end}). 
Policies are implemented as LSTM actor-critics trained over complete $H{=}35$-year episodes, with the hidden state carried forward through time.

\paragraph{Policy Scenarios}
We evaluate two versions of the climate-policy game to compare surrogate and simulator performance:

\emph{(i) Tractable scenario.} $N=4$ homogeneous agents (identical damage and cost parameters). 
Only the energy decarbonization lever has a meaningful effect (the other actions are tuned to be prohibitively costly or yield negligible climate impact). 
This scenario provides strong feedback signals and converges quickly, allowing us to test whether the surrogate-based environment yields the same policies as the original simulator (parameters in Appendix~D).

\emph{(ii) Intractable scenario.} $N=10$ heterogeneous agents with diverse emission shares, damage sensitivities, and mitigation costs. 
Multiple mitigation levers produce similar climate outcomes, making it difficult for agents to discern which actions are most effective (resulting in weak learning signals). 
As a result, this scenario requires vastly more environment interactions to learn good policies.
Training to convergence with the full CICERO-SCM model would be prohibitively slow (millions of simulator calls), so we use the surrogate to improve the computational scalability of the climate-transition component in this scenario (parameters in Appendix~D).

\subsection{Evaluation Criteria}
\label{sec:eval_metrics}
We evaluate the surrogate $f_\mathrm{NET}$ against $f_{\mathrm{SCM}}$) across four dimensions: accuracy, inference speed, training acceleration, and policy consistency. 

\paragraph{Accuracy}
We compute RMSE and coefficient of determination $R^2$ between predicted and true temperature increments $\Delta \hat T(t)$ and $\Delta T(t)$ on a held-out test set.

\paragraph{Inference Time}
Both models are wrapped in classes initialized with historical emissions and a step method that appends new emissions $E(t)$, performs preprocessing, and returns $\Delta T(t)$ from inference computation. 
We measure the mean runtime over $100{,}000$ steps, with CICERO-SCM run on CPU and the RNN surrogate on both CPU and GPU.

\paragraph{MARL Time}
We compare per environment step time under two otherwise identical environments (scenario (i)) that differ only in the climate backend. 
CICERO-SCM is executed on CPU whereas the RNN-based surrogate is executed on GPU.

\paragraph{Policy Consistency}
An ideal surrogate should induce policies indistinguishable from those obtained with the original simulator. 
Let $\Pi$ denote the policy class and $J_f(\pi)=\mathbb{E}\!\left[\sum_{t=0}^{H}\gamma^{t} r_t \mid f,\pi\right]$ denote the expected discounted return under climate engine $f$ and policy $\pi \in \Pi$.
We use two criteria to motivate policy consistency:
\begin{align}
\operatorname{sign}\!\big[\Delta J_{f_{\mathrm{NET}}}(\pi_1,\pi_2)\big]
&\approx
\operatorname{sign}\!\big[\Delta J_{f_{\mathrm{SCM}}}(\pi_1,\pi_2)\big]
\label{eq:pref_order}
\\
\nabla_{\theta} J_{f_{\mathrm{NET}}}(\pi_\theta)
&\approx
\nabla_{\theta} J_{f_{\mathrm{SCM}}}(\pi_\theta)
\label{eq:gradient}
\end{align}
where $\Delta J_f(\pi_1,\pi_2)=J_f(\pi_1)-J_f(\pi_2)$ is the reward difference between two policies $\pi_1,\pi_2\in\Pi$ and $\theta\in\mathcal{N}(\theta^\star_{\mathrm{SCM}})$ is the set of parameters in the policy-induced visitation distribution under $f_{\mathrm{SCM}}$. 
This definition is conceptually related to model-based RL analyses of return and gradient alignment between learned dynamics and true environments \cite{janner2021_whentotrust}. In model-based RL terms, eq.~(\ref{eq:pref_order}) relates to preserving policy value ordering, while eq.~(\ref{eq:gradient}) relates to matching local improvement directions, which depend on surrogate model error along the policy-induced visitation distribution.
This is consistent with analyses such as \cite{adaptation_po}, which suggest that controlling model error on a policy’s visitation distribution controls return discrepancies and helps preserve policy rankings and local improvement directions. However, directly validating eqs.~(\ref{eq:pref_order})–(\ref{eq:gradient}) would require training with $f_{\mathrm{SCM}}$, which is intractable in our large-scale setting.

Hence, we propose a tractable empirical alternative. Let's first define the feasible set of emissions trajectories attainable in our MARL game $\mathcal{K}$ as:
\begin{align}
\begin{aligned}
\mathcal K 
&:= \bigl\{\, \bar E(\cdot)
\ \bigm|\ 
\bar E(t), \\
&\qquad
\tilde\delta_{\mathcal{G}\setminus\mathcal{C}}(t) = 0, \quad
\tilde\delta_{\mathcal{C}}(t) \in \mathcal D(M,\mathcal L)
\,\bigr\}
\end{aligned}
\end{align}
where the link between $\tilde\delta$ and $\bar E$ is described in Section~\ref{sec:marl}, and $\mathcal{D}$ is the set of controllable growth deviations induced by the lever levels $k_{i,t} \in \mathcal{L}$ and the policy matrix $M$.

Let $\mathcal{S} \subseteq \mathcal{K}$ be defined as the emission trajectories stored during training using $f_{\mathrm{NET}}$ more formally defined as:
\begin{align}
\mathcal{S} \;&=\; \{\bar E^{(k)}(\cdot)\}_{k=1}^K \\
\bar E^{(k)}(\cdot) &= \big(\bar E^{(k)}(1),\dots,\bar E^{(k)}(H+U)\big)
\end{align}
where $\bar E^{(k)}(t)\in\mathbb{R}^{|\mathcal{G}|}$, $K$ denotes the total number of training episodes, $H$ is the episode length, and $U$ is the rollout length.
Then let $\bar{\mathcal S}\subseteq\mathcal K$ denote the set of emission trajectories that would be stored during training using $f_{\mathrm{SCM}}$ if that would have been tractable.
Trivially, $\mathcal{S} = \bar{\mathcal S}$ if $f_{\mathrm{NET}} = f_{\mathrm{SCM}}$ (all else equal).

More generally, if the surrogate uniformly approximates the simulator on $\mathcal K$:
\begin{align}
\sup_{\bar E(\cdot)\in\mathcal{K}}
\big\|f_{\mathrm{NET}}(\bar E(\cdot)) - f_{\mathrm{SCM}}(\bar E(\cdot))\big\|_\infty
\;\le\; \varepsilon \label{eq:sup_diff}
\end{align}
then for sufficiently small $\varepsilon$ one expects the sets of policy-induced trajectories to satisfy:
\begin{align}
\big\|\mathcal S - \bar{\mathcal S}\big\| \;\to\; 0 
\quad \text{as} \quad \varepsilon \to 0. \label{eq:s_s}
\end{align}
While we do not prove this formally, and therefore do not claim policy equivalence, the intuition is that a surrogate that accurately approximates the simulator will have policy-induced emission trajectories $\mathcal S$ that are arbitrarily close to those of the true simulator $\bar{\mathcal S}$. Moreover, because MARL exploration introduces stochasticity in the agents’ policy updates, the sampled trajectories may already overlap substantially with $\bar{\mathcal S}$ even for moderate approximation errors ($\varepsilon>0$).

Therefore, we propose to randomly sample $N$ trajectories from $\mathcal{S}$ (uniform without replacement) and replay them through $f_{\mathrm{SCM}}$ to obtain $\Delta T^{\mathrm{SCM}}(\cdot)$. 
While $\mathcal{S}\neq\bar{\mathcal S}$ in general, a sufficiently accurate surrogate implies substantial overlap between $\mathcal{S}$ and $\bar{\mathcal S}$, so samples from $\mathcal{S}$ provide a reasonable proxy for the policy-induced emission trajectory distribution under the true simulator $\bar{\mathcal S}$. 
This intuition mirrors the transition-occupancy-matching approach presented in \cite{policyaware_mbrl}, which learns a dynamics model by matching the distribution of transitions experienced by the current policy in the real environment and in the model.
By focusing on policies along the convergence trajectory rather than random states, we evaluate the surrogate in the regions of the state-action space most relevant for converging to the same policies.

Consequently, computing RMSE between $\Delta T^{\mathrm{NET}}(\cdot)$ and $\Delta T^{\mathrm{SCM}}(\cdot)$ for $N$ sampled trajectories in $\mathcal{S}$ provides an empirical proxy for the surrogate's accuracy on policy-induced emission paths. 
As a proxy for preservation of preference ordering (eq. \ref{eq:pref_order}), we also compute Kendall's $\tau$ rank-consistency between discounted temperature-based returns defined as:
\begin{align}
    r^f_k =-\sum_t \gamma^t \Delta T^{f}_k(t), \quad f \in\{f_{\mathrm{SCM}}, f_{\mathrm{NET}} \}
\end{align}
where $k=[1,\dots,N]$.
Evaluating only $N\!\ll\!|\mathcal{S}|$ trajectories keeps the evaluation tractable when training $\pi^{\mathrm{SCM}}$ to convergence is intractable.

\begin{table*}[t]
  \centering
  \resizebox{\textwidth}{!}{%
    \renewcommand{\arraystretch}{1.1}
    \setlength{\tabcolsep}{4pt}
    \begin{tabular}{l ccc cccc}
      \toprule
        & \multicolumn{3}{c}{\textbf{Test data performance \& inference speed}} 
        & \multicolumn{4}{c}{\textbf{Policy-induced performance \& MARL speed}} \\
        \cmidrule(lr){2-4} \cmidrule(lr){5-8}
        \textbf{Climate engine} 
        & \textbf{\shortstack[c]{Test data\\(RMSE, $R^2$)}} 
        & \textbf{\shortstack[c]{Mean inference [s]\\(CPU/GPU)}} 
        & \textbf{\shortstack[c]{Speed-up\\(CPU/GPU)}} 
        & \textbf{\shortstack[c]{Scenario (i) \\(RMSE, rank-$\tau$)}} 
        & \textbf{\shortstack[c]{Scenario (ii) \\(RMSE, rank-$\tau$)}} 
        & \textbf{\shortstack[c]{Mean\\ env-step [s]}} 
        & \textbf{Speed-up} \\
      \midrule
      CICERO\textendash SCM  
        & -- & $464.4\times 10^{-3}\,$ / -- & -- 
        & -- & -- & $217.7\times 10^{-3}\,$ & --\\
      LSTM  
        & ($4.7\times 10^{-4}\,$, 0.99) & $1.1\times 10^{-3}\,$ / $0.4\times 10^{-3}\,$ & $442{\times}$ / $1161{\times}$
        & ($5.9\times 10^{-4}\,$, 0.996) & ($3.2\times 10^{-4}\,$, 0.990) & $1.6\times 10^{-3}\,$ & $137{\times}$\\
      GRU  
        & ($3.7\times 10^{-4}\,$, 0.99) & $2.3\times 10^{-3}\,$ / $0.4\times 10^{-3}\,$ & $202{\times}$ / $1161{\times}$
        & ($3.9\times 10^{-4}\,$, 0.996) & ($2.0\times 10^{-4}\,$, 0.997) & $1.6\times 10^{-3}\,$ & $137{\times}$\\
      TCN  
        & ($6.8\times 10^{-4}\,$, 0.99) & $3.3\times 10^{-3}\,$ / $1.3\times 10^{-3}\,$ & $140{\times}$ / $357{\times}$
        & ($21.1\times 10^{-4}\,$, 0.994) & ($10.3\times 10^{-4}\,$, 0.982) & $4.5\times 10^{-3}\,$ & $49{\times}$\\
      \bottomrule
    \end{tabular}
  }
  \caption{Surrogate performance on held-out test data and policy-induced trajectories and acceleration of inference speed and MARL environment step in scenario (i). Speed-up is measured relative to CICERO-SCM.}
  \label{tab:results_table}
\end{table*}

\begin{figure}[t]
  \centering
  \includegraphics[width=0.95\linewidth]{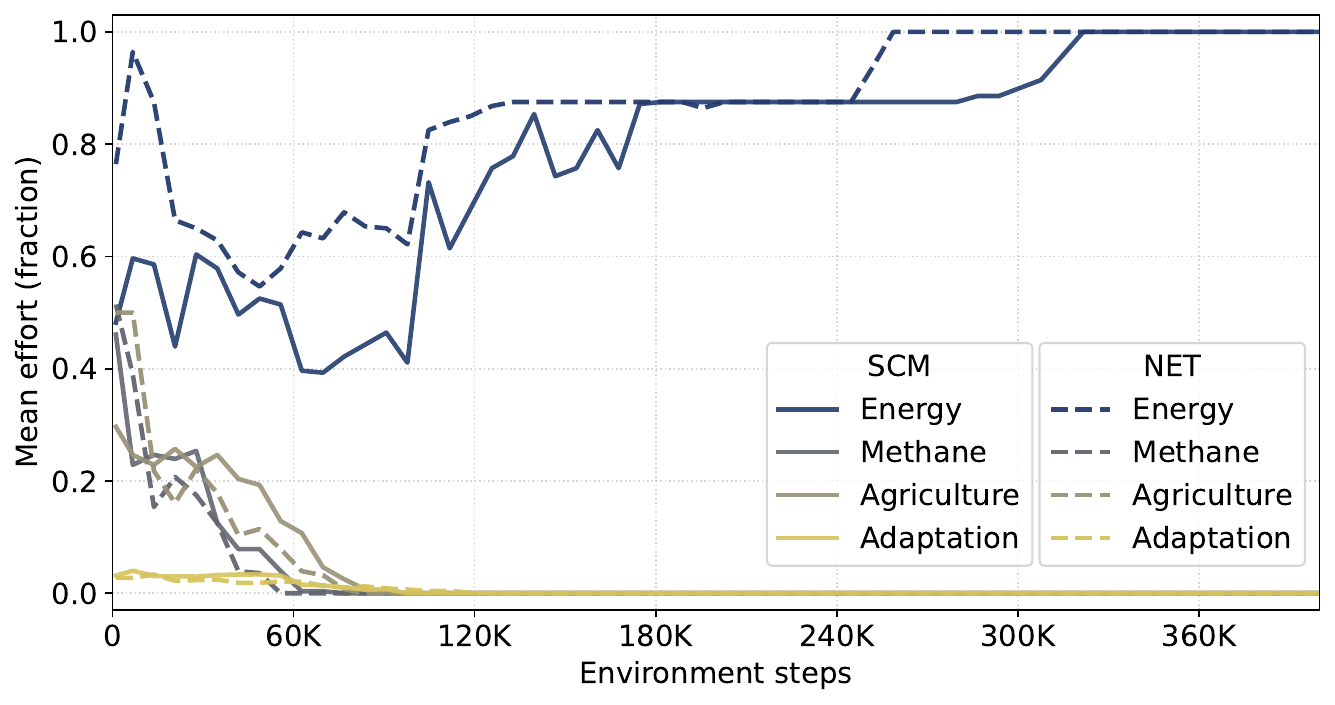}
    \caption{Comparison of learned policies in tractable scenario (i) between CICERO-SCM and the GRU-based surrogate.}
  \label{fig:learned_policy_consistency}
\end{figure}

\section{Results} \label{sec:results}
\subsection{Surrogate Model Performance}
The left side of Table~\ref{tab:results_table} summaries the surrogate accuracy and computational efficiency on held‑out data.
All three surrogates achieve nearly perfect accuracy, with GRU being most accurate (RMSE $3.7\times 10^{-4}\,$K) and TCN least (RMSE $6.8\times 10^{-4}\,$K).
Given that the training trajectories span $0$–$2.5$K (Figure \ref{fig:temp_ensemble}), these errors are very small. 
LSTM and GRU variants have the fastest one-step inference on GPU ($\sim\!0.0004$s), translating to $\sim\!1100\times$ faster than CICERO‑SCM.
This makes embedding climate dynamics in MARL feasible, where simulator use would be intractable.
The TCN is slower, requiring 5 layers to reach $10^{-4}$ accuracy, whereas LSTM and GRU needed only 1 layer.
The one‑step speedups translate into $>\!100\times$ faster per-environment call time during MARL training.
Speed-up is not linear with inference time, as policy networks and communication overhead begin to dominate once the surrogate becomes fast.
We run 32 environments per runner, further hiding step-function latency.
Thus, the comparison is conservative - the speed-up would appear larger without parallelism.

\subsection{Evaluating Policy Consistency}
High test-set accuracy does not necessarily mean that the surrogate will induce the same policies as CICERO-SCM.
Agents trained with different climate engines may visit different state trajectories, and small prediction errors can compound over long horizons.
To assess whether policies learned with $f_{\mathrm{NET}}$ remain consistent with those that would emerge under $f_{\mathrm{SCM}}$, we replay $N\!=\!1000$ emission trajectories sampled from the policy-induced distribution visited during training.
For each trajectory, we compare the resulting temperature paths from the surrogate and simulator using the RMSE and the Kendall’s $\tau$ rank-consistency between discounted temperature-based returns.
These metrics quantify accuracy along realistic, policy-relevant trajectories and whether the surrogate preserves the preference ordering over policies implied by the simulator.

\paragraph{(i) Tractable Scenario}
In the homogeneous setting, both surrogate and simulator can be trained to convergence.
Table~\ref{tab:results_table} shows that LSTM and GRU maintain low RMSE, and Kendall’s $\tau$ confirms return rankings are preserved.
These results support the policy consistency criteria from eqs.~(\ref{eq:pref_order}–\ref{eq:gradient}).
Figure~\ref{fig:learned_policy_consistency} shows that, in scenario (i) with only the energy-decarbonization lever activated, the mean lever activation (mean across agents) converges to the same trajectory under both $f_\text{NET}$ and $f_\text{SCM}$ (more in Appendix~D, Figure~A.8).

\paragraph{(ii) Intractable Scenario}
In the heterogeneous $N\!=\!10$ case, convergence is intractable using the simulator, so we train with $f_{\mathrm{NET}}$ for $>\!1$M steps.
As shown in Table~\ref{tab:results_table}, replayed policy-induced trajectories yield low RMSE and high Kendall’s $\tau$, indicating consistent policy ranking under $f_{\mathrm{SCM}}$.
A likely reason for the strong replay metrics is that each agent controls a smaller emissions share, leading to smoother, less extreme deviations from SSP2-4.5 and trajectories that remain closer to the surrogate’s training distribution.
While replay does not provide a formal guarantee, it indicates small discrepancies on the policy-relevant region and preserves the return ordering across policies.
Qualitatively, some agents converge to near-max methane abatement, others maintain moderate energy decarbonization, agriculture is used strongly by a subset of agents, while adaptation remains close to zero (Appendix~D, Fig.~A.8).

\section{Limitations and Future Research} \label{sec:limitations}
The results presented in this paper demonstrate the benefits of using surrogate climate models within climate-economic MARL settings, however, there are limitations and directions for future research.

\emph{Refinement of MARL Experiment.} More complex experimental setups could incorporate cooperation mechanisms and heterogeneous damage functions, preferably via surrogates of local high-fidelity simulators - an approach our climate collaborators have expressed interest in.
Such extensions would primarily modify the climate-impact interface, rather than the MARL loop itself.

\emph{Uncertainty-aware Surrogates.} We did not perturb the calibration parameters of CICERO-SCM’s differential-equation core when training the RNN surrogates. 
Conditioning the surrogate on these parameters, and potentially predicting distributions rather than point estimates, would propagate structural uncertainty through the MARL loop and enable distributional or risk-sensitive objectives \cite{bellemare2017distributional}, which are relevant under tipping-point and tail-risk scenarios.

\emph{Policy Consistency.} While we propose an empirical method to test policy consistency, a formal proof of eqs. (\ref{eq:sup_diff})–(\ref{eq:s_s}) would further substantiate the claim. Our conclusions do not depend on such a proof, and we leave it for future work.

\section{Conclusion} \label{sec:conclusions}
In collaboration with the European Environment Agency, we introduced a framework for integrating higher-fidelity climate dynamics into scalable multi-agent reinforcement learning by replacing the climate module with a learned surrogate.
We developed an RNN-based emulator of the CICERO-SCM climate model trained on $20{,}000$ multi-gas emission trajectories, achieving global-mean temperature RMSE $<\!4\times 10^{-4}$K and $\sim\!1000\times$ faster one-step inference, translating into $>\!100\times$ end-to-end MARL training speed-ups.

Using the surrogate within MARL, we converge to the same set of policies in a computationally tractable experiment; when direct validation is infeasible, we propose replaying policy-induced emission trajectories through the simulator as a tractable consistency check.
Together, these results remove a major computational barrier to scalable research on cooperative climate-policy design and uncertainty propagation.

\section*{Acknowledgments}
The work presented in this article is supported by Novo Nordisk Foundation grant NNF23OC0085356.

\bibliographystyle{named}
\bibliography{ijcai26}

\clearpage

\onecolumn
\appendix
\renewcommand{\thetable}{A.\arabic{table}}
\renewcommand{\thefigure}{A.\arabic{figure}}
\setcounter{table}{0}
\setcounter{figure}{0}

\section*{Code availability}
An implementation of all surrogate models and MARL experiments is publicly available at:
\url{https://github.com/oskarbohnlassen/ciceroscm-surrogate}.

\section{CICERO-SCM}
\label{app:cicero-scm}
We use CICERO–SCM (v1.1.1, Python) as the reference climate engine that maps multi-gas emissions to global-mean temperature change. This appendix documents the concrete inputs and parameterization we used so that the experiments in Section \ref{sec:cicero} are fully reproducible. 
We keep the model as a black box in the main text but here we expose the file layout, key parameter groups, and the exact calibration we use.

In CICERO-SCM v1.1.1, inputs are organized into structured files that define the model species and reference pathways. 
The file \texttt{gases\_v1RCMIP.txt} lists all species (simplified version in Table~\ref{tab:gases}) together with their decay and forcing parameters. 
Emission and concentration time series are provided in RCMIP format (e.g., \texttt{ssp245\_conc\_RCMIP.txt}, \texttt{ssp245\_em\_RCMIP.txt}), where the files here represent SSP2-4.5, a ''middle-of-the-road'' socio-economic scenario with $4.5\ \mathrm{Wm^{-2}}$ radiative forcing in 2100.
Natural background emissions (e.g., \texttt{natemis\_ch4.txt}, \texttt{natemis\_n2o.txt}) are specified separately.

We also list the internal CICERO-SCM naming conventions corresponding to the parameter groups specified in Table~\ref{tab:scm_params}.
Climate response parameters correspond to the CICERO group \texttt{pamset\_udm}, which includes coefficients governing the upwelling-diffusion energy balance model (air-sea heat exchange, ocean diffusivity and upwelling, mixed-layer heat capacity, polar amplification, and interhemispheric heat exchange).
Emissions-to-forcing parameters correspond to the CICERO group \texttt{pamset\_emiconc}, which specifies scaling factors for converting emissions of species such as SO$_2$, ozone, black carbon, and organic carbon into concentrations and effective radiative forcing.
The baseline parameter values used in this study are drawn from CICERO’s official calibration suite and correspond to the configuration distributed under the name \texttt{13555\_old\_NR\_improved}.

\begin{table*}[ht]
\centering
\renewcommand{\arraystretch}{1.05}
\begin{tabular}{llll}
\toprule
\textbf{Species} & \textbf{Class} & \textbf{Forcing Sign} & \textbf{Model Treatment} \\
\midrule
CO\textsubscript{2} (FF)      & Long-lived GHG            & Warming & Carbon-cycle \\
CO\textsubscript{2} (AFOLU)   & Long-lived GHG            & Warming & Carbon-cycle \\
CH\textsubscript{4}           & Short-lived GHG           & Warming & Simplified decay (multi-$\tau$) \\
N\textsubscript{2}O           & Long-lived GHG            & Warming & Fixed lifetime decay \\
SO\textsubscript{2}           & Aerosol precursor         & Cooling & Linear forcing proxy \\
CFC-11                        & Long-lived GHG            & Warming & Fixed lifetime decay \\
CFC-12                        & Long-lived GHG            & Warming & Fixed lifetime decay \\
CFC-113                       & Long-lived GHG            & Warming & Fixed lifetime decay \\
CFC-114                       & Long-lived GHG            & Warming & Fixed lifetime decay \\
CFC-115                       & Very-long-lived GHG            & Warming & Fixed lifetime decay \\
CH\textsubscript{3}Br         & Short-lived GHG            & Warming & Fixed lifetime decay \\
CCl\textsubscript{4}          & Long-lived GHG            & Warming & Fixed lifetime decay \\
CH\textsubscript{3}CCl\textsubscript{3} & Short-lived GHG  & Warming & Fixed lifetime decay \\
HCFC-22                       & Long-lived GHG            & Warming & Fixed lifetime decay \\
HCFC-141b                     & Short-lived GHG            & Warming & Fixed lifetime decay \\
HCFC-123                      & Short-lived GHG            & Warming & Fixed lifetime decay \\
HCFC-142b                     & Long-lived GHG            & Warming & Fixed lifetime decay \\
H-1211                        & Long-lived GHG            & Warming & Fixed lifetime decay \\
H-1301                        & Long-lived GHG            & Warming & Fixed lifetime decay \\
H-2402                        & Long-lived GHG            & Warming & Fixed lifetime decay \\
HFC-125                       & Long-lived GHG            & Warming & Fixed lifetime decay \\
HFC-134a                      & Long-lived GHG            & Warming & Fixed lifetime decay \\
HFC-143a                      & Long-lived GHG            & Warming & Fixed lifetime decay \\
HFC-227ea                     & Long-lived GHG            & Warming & Fixed lifetime decay \\
HFC-23                        & Very-long-lived GHG            & Warming & Fixed lifetime decay \\
HFC-245fa                     & Short-lived GHG            & Warming & Fixed lifetime decay \\
HFC-32                        & Short-lived GHG            & Warming & Fixed lifetime decay \\
HFC-4310mee                   & Long-lived GHG            & Warming & Fixed lifetime decay \\
C\textsubscript{2}F\textsubscript{6} & Very-long-lived GHG & Warming & Fixed lifetime decay \\
C\textsubscript{6}F\textsubscript{14} & Very-long-lived GHG & Warming & Fixed lifetime decay \\
CF\textsubscript{4}           & Very-long-lived GHG       & Warming & Fixed lifetime decay \\
SF\textsubscript{6}           & Very-long-lived GHG       & Warming & Fixed lifetime decay \\
NO\textsubscript{x}           & Ozone precursor           & Mixed   & Linear forcing proxy \\
CO                            & Ozone precursor           & Warming & Linear forcing proxy \\
NMVOC                         & Ozone precursor           & Mixed   & Linear forcing proxy \\
NH\textsubscript{3}           & Aerosol precursor         & Cooling & Linear forcing proxy \\
BMB\_AEROS\_BC                & Aerosol precursor         & Warming & Linear forcing proxy \\
BMB\_AEROS\_OC                & Aerosol precursor         & Cooling & Linear forcing proxy \\
BC                            & Aerosol precursor         & Warming & Linear forcing proxy \\
OC                            & Aerosol precursor         & Cooling & Linear forcing proxy \\
\bottomrule
\\
\end{tabular}
\caption{CICERO-SCM species grouped by class and defined by atmospheric lifetime, forcing sign, and model treatment.}
\label{tab:gases}
\end{table*}

\begin{table*}[ht]
  \centering
  \renewcommand{\arraystretch}{1.2}
  \setlength{\tabcolsep}{6pt}
  \begin{tabular}{p{2.5cm} l c l}
    \toprule
    \textbf{Group} & \textbf{Parameter} & \textbf{Value} & \textbf{Description} \\
    \midrule
    \multirow{7}{=}{\centering Climate response \\ parameters} 
      & rlamdo & 15.0836  & Air–sea heat exchange parameter [W\,m$^{-2}$\,K$^{-1}$] \\
      & akapa  & 0.6568   & Vertical heat diffusivity [cm$^{2}$\,s$^{-1}$] \\
      & cpi    & 0.2077   & Polar amplification factor \\
      & W      & 2.2059   & Upwelling velocity [m\,yr$^{-1}$] \\
      & beto   & 6.8982   & Ocean heat exchange coefficient [W\,m$^{-2}$\,K$^{-1}$] \\
      & lambda & 0.6063   & Climate sensitivity parameter [K\,W$^{-1}$\,m$^{2}$] \\
      & mixed  & 107.2422 & Mixed-layer ocean depth [m] \\
    \midrule
    \multirow{6}{=}{\centering Emissions-to-\\forcing parameters} 
      & qbmb    & 0.0      & Biomass burning forcing coefficient \\
      & qo3     & 0.5      & Tropospheric ozone forcing coefficient \\
      & qdirso2 & -0.3562  & Direct SO$_2$ forcing coefficient \\
      & qindso2 & -0.9661  & Indirect SO$_2$ forcing coefficient \\
      & qbc     & 0.1566   & Black carbon forcing coefficient \\
      & qoc     & -0.0806  & Organic carbon forcing coefficient \\
    \bottomrule
    \\
  \end{tabular}
  \caption{Baseline parameter configuration used in CICERO-SCM. Parameters are grouped into those governing the climate system response (top) and those scaling emissions into concentrations and effective radiative forcing (bottom).}
  \label{tab:scm_params}
\end{table*}

\newpage
\FloatBarrier

\section{Generated emission trajectories}
\begin{figure*}[h]
  \centering
  \begin{subfigure}{0.45\textwidth}
    \centering
    \includegraphics[width=0.95\linewidth]{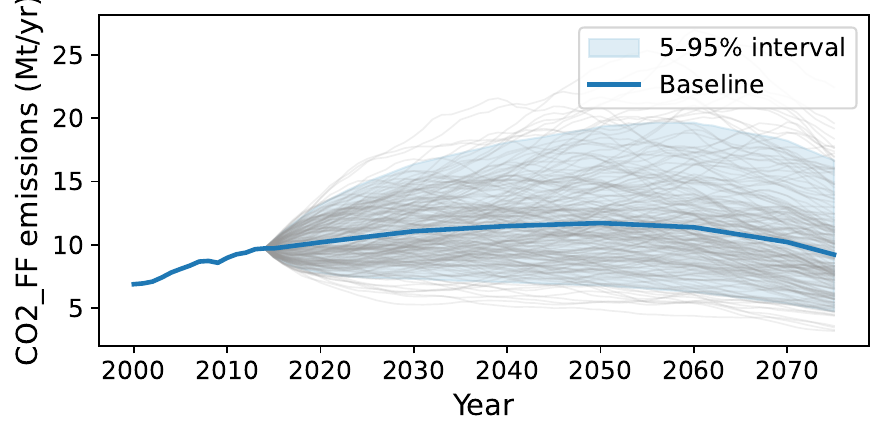}
    \caption{CO$_2$ (FF)}
  \end{subfigure}
  \hfill
  \begin{subfigure}{0.45\textwidth}
    \centering
    \includegraphics[width=0.95\linewidth]{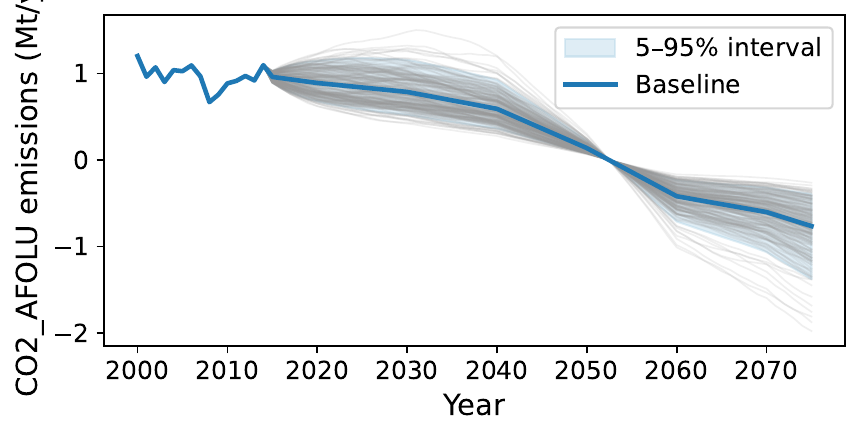}
    \caption{CO$_2$ (AFOLU)}
  \end{subfigure}

  \vspace{0.5em}
  \begin{subfigure}{0.45\textwidth}
    \centering
    \includegraphics[width=0.95\linewidth]{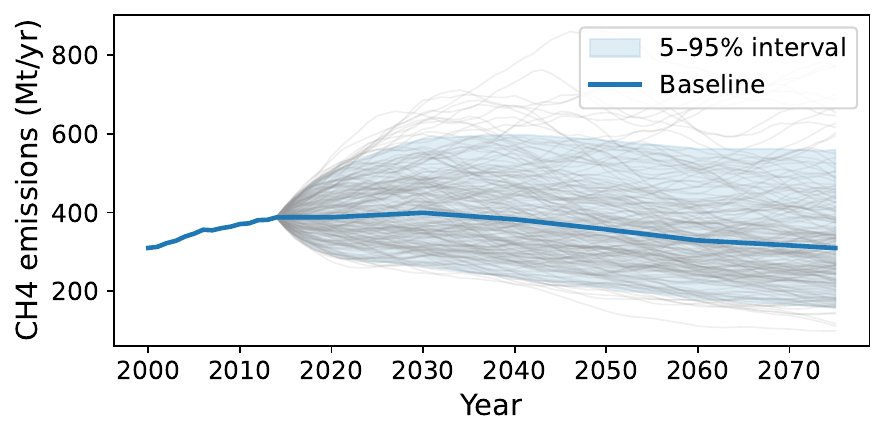}
    \caption{CH$_4$}
  \end{subfigure}
  \hfill
  \begin{subfigure}{0.45\textwidth}
    \centering
    \includegraphics[width=0.95\linewidth]{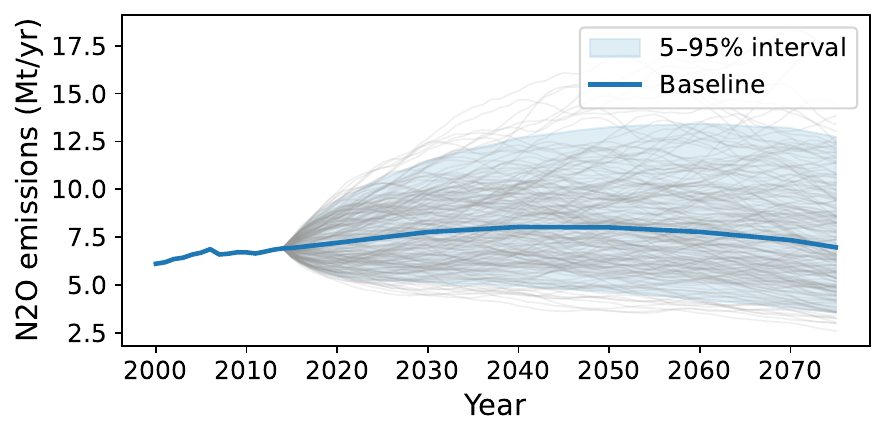}
    \caption{N$_2$O}
  \end{subfigure}

  \vspace{0.5em}
  \begin{subfigure}{0.45\textwidth}
    \centering
    \includegraphics[width=0.95\linewidth]{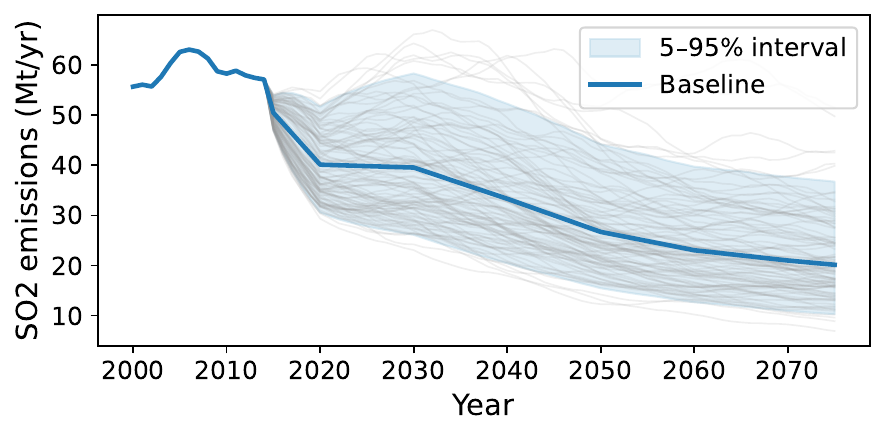}
    \caption{SO$_2$}
  \end{subfigure}
  \caption{Ensemble of generated emission trajectories for the five controllable gases. Shaded regions represent the 5–95\% range across the 20,000 generated scenarios, solid lines indicate the ensemble median, dashed lines mark the SSP2-4.5 baseline.}
  \label{fig:app_emissions}
\end{figure*}

In Section \ref{sec:surrogate} we generate an ensemble of policy-relevant multi-gas trajectories by perturbing the SSP2-4.5 baseline growth.
Figure~\ref{fig:app_emissions} illustrates an ensemble of emission trajectories for the five controllable gases in $\mathcal{C}$ over 2015-2075.

\newpage
\FloatBarrier

\section{Surrogate models}
\begin{figure*}[ht]
  \centering
  
  \begin{subfigure}{0.6\textwidth}
    \centering
    \includegraphics[width=0.9\linewidth]{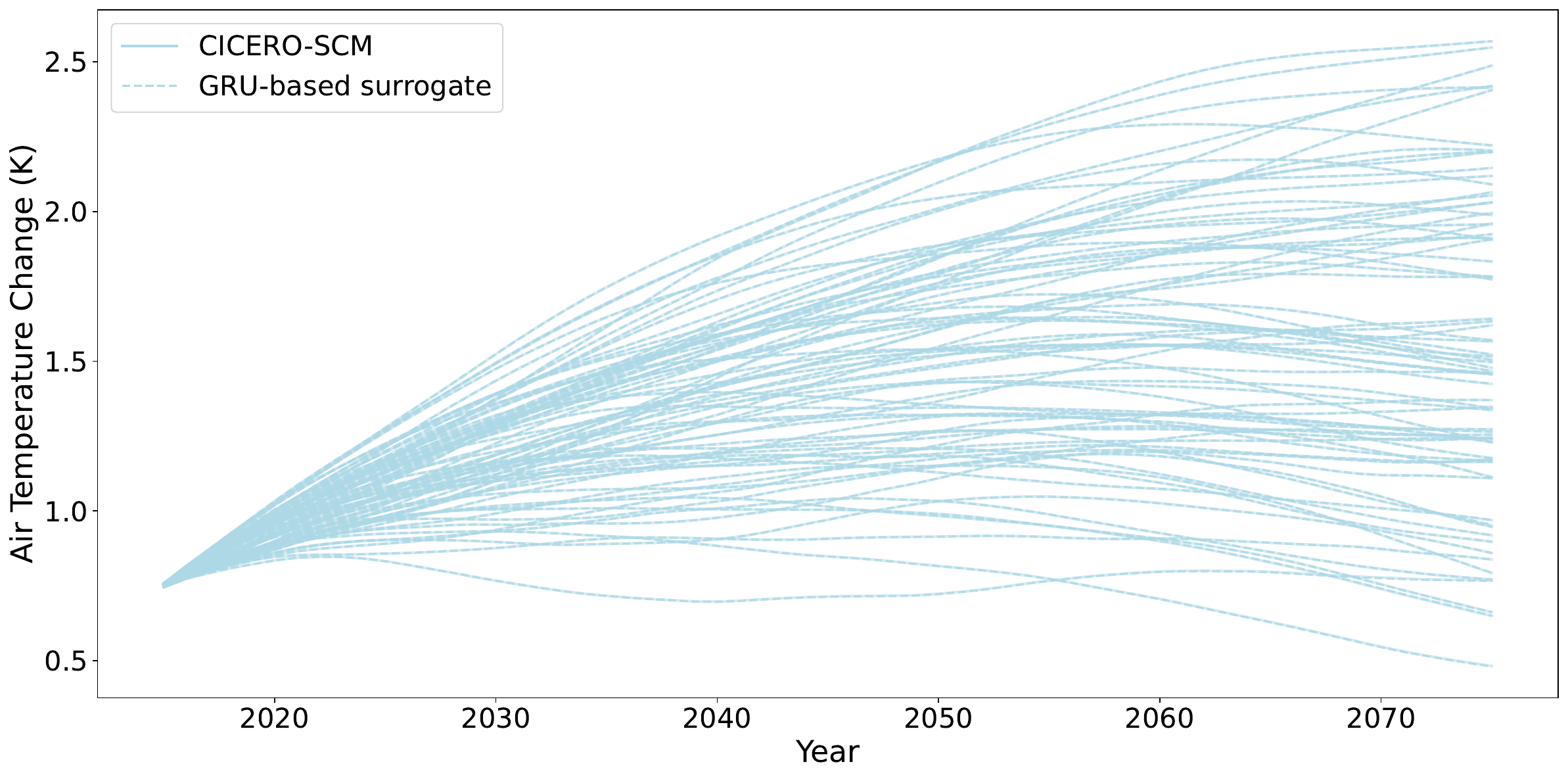}
    \caption{GRU}
  \end{subfigure}
  
  \vspace{0.5em}

  \begin{subfigure}{0.6\textwidth}
    \centering
    \includegraphics[width=0.9\linewidth]{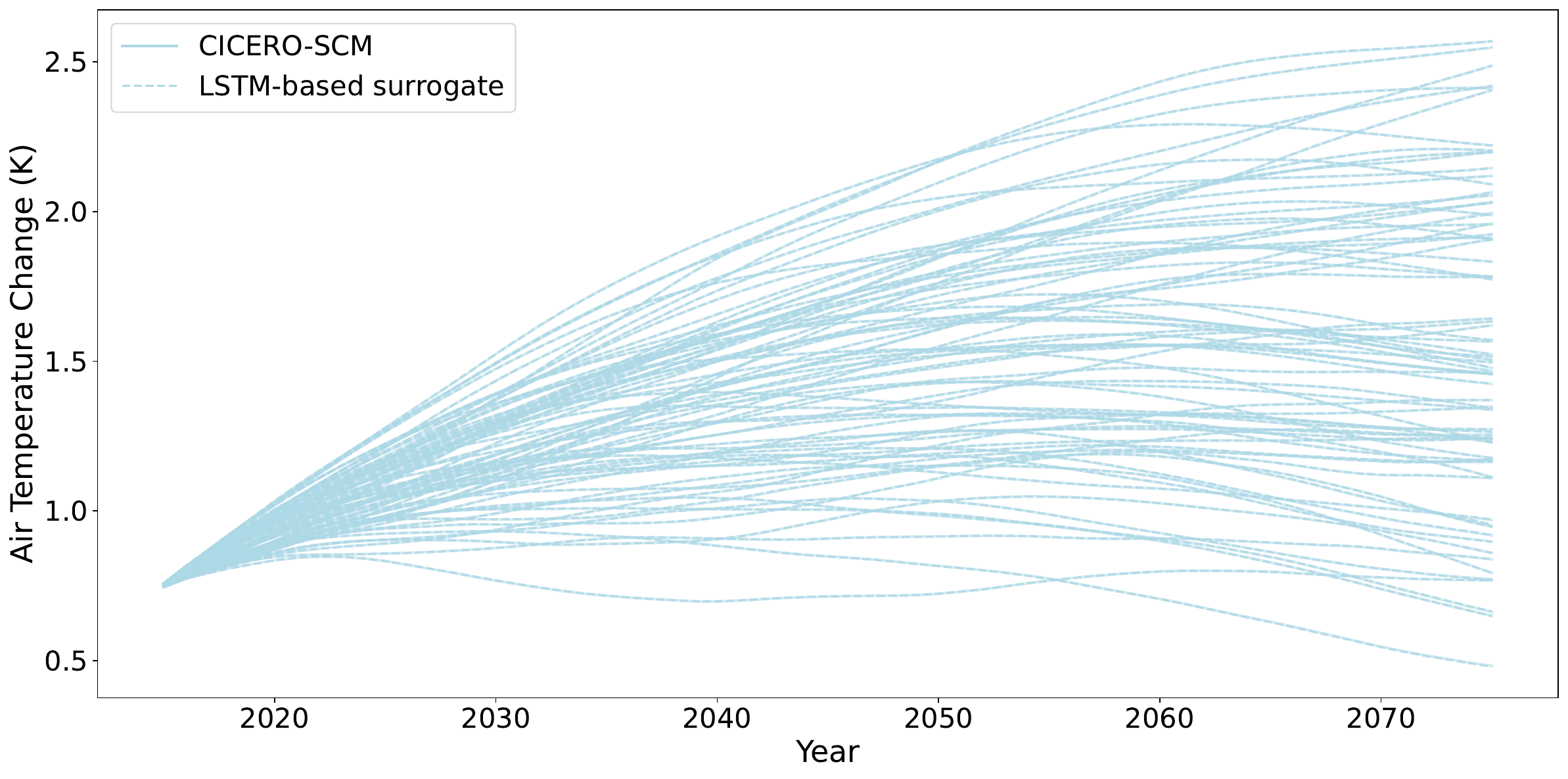}
    \caption{LSTM}
  \end{subfigure}

  \vspace{0.5em}
  
  \begin{subfigure}{0.6\textwidth}
    \centering
    \includegraphics[width=0.9\linewidth]{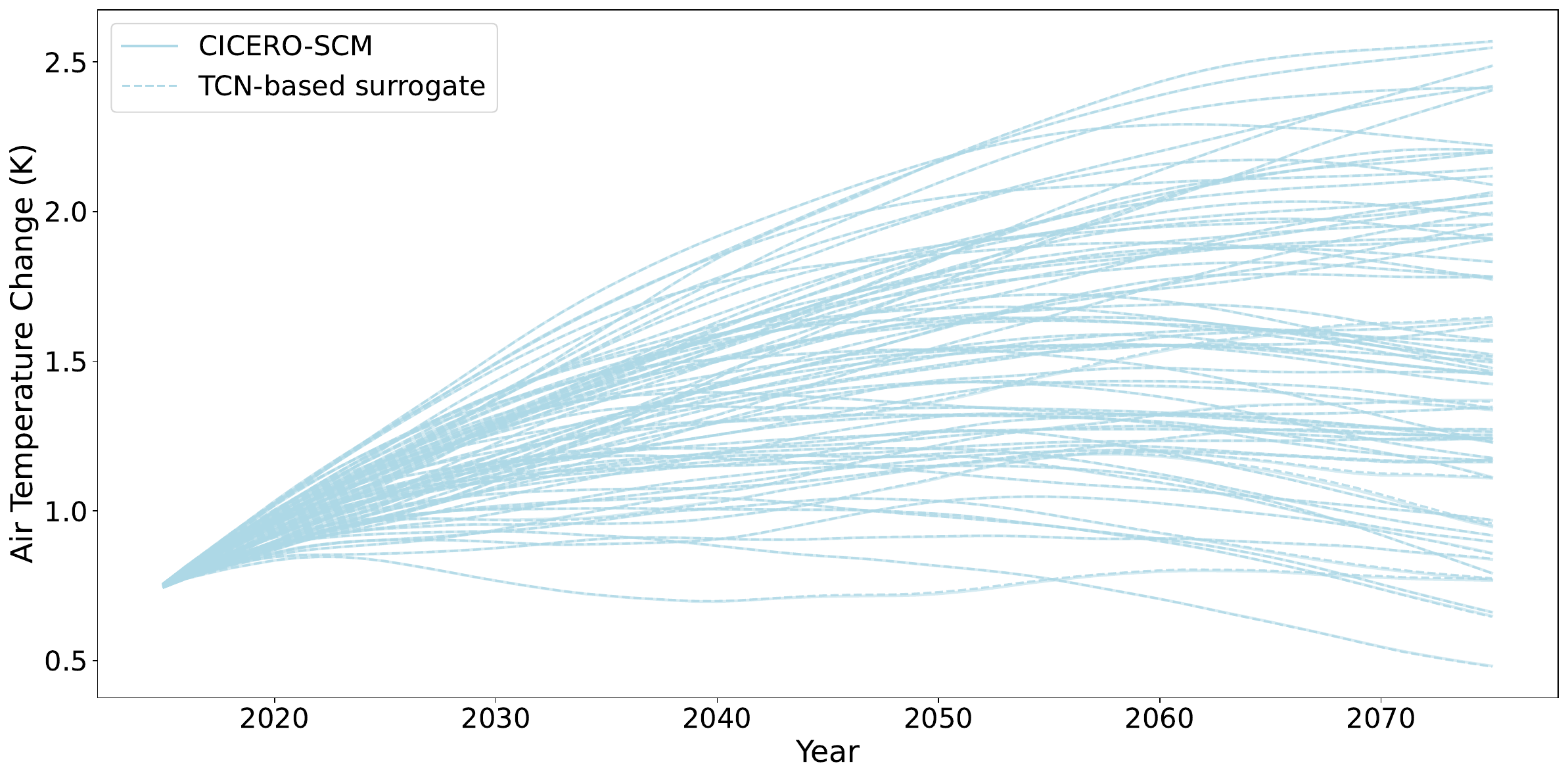}
    \caption{TCN}
  \end{subfigure}
  
  \caption{Comparison of temperature trajectories simulated by CICERO-SCM (solid) and the RNN-based surrogates (dashed) for randomly selected sequences in the test data. 
  Each panel shows results for one architecture (GRU, LSTM, and TCN), illustrating the perfect agreement between surrogate predictions and the ground-truth simulator.}
  \label{fig:test_seq}
\end{figure*}

The performance metrics of the RNN-based surrogates were presented in Section \ref{sec:results}. 
Figure \ref{fig:test_seq} illustrate how the impressively low RMSE leads to what looks to be perfect agreement between the surrogates and the simulator.

\newpage
\FloatBarrier

\section{MARL experiment}
\label{app:marl_experiments}
We provide implementation details for the two scenarios described in Section \ref{sec:marl}. Table \ref{tab:marl_env_scenario_i} shows the parameters used for scenario (i) whereas Table \ref{tab:marl_env_scenario_ii} shows the parameters used for scenario (ii).

\begin{table*}[h]
  \centering
  \small
  \renewcommand{\arraystretch}{1.15}
  \setlength{\tabcolsep}{6pt}
  \begin{tabular}{p{3.2cm} p{12.2cm}}
    \toprule
    \textbf{Item} & \textbf{Specification} \\
    \midrule
    Agents ($N$) & $N=4$ agents with shares $S_i = [0.25,\,0.25,\,0.25,\,0.25]$ \\
    Climate engines & CICERO-SCM and surrogates \\
    Controlled gases & CO2\_FF, CO2\_AFOLU, CH4, N2O, SO2 \\
    \midrule
    \multicolumn{2}{l}{\textbf{Levers and levels}} \\
    Energy & Levels: $0.0,\ 0.5,\ 1.0$ \\
    Methane & Levels: $0.0,\ 0.5,\ 1.0$ \\
    Agriculture & Levels: $0.0,\ 0.5,\ 1.0$ \\
    Adaptation & Levels: $0.00,\ 0.03,\ 0.08$ \\
    \midrule
    \multicolumn{2}{l}{\textbf{Policy to emissions mapping (per-year $\Delta$ growth coefficients)}} \\
    Energy & CO2\_FF: $-0.05$, CH4: $-0.005$, N2O: $-0.005$, SO2: $-0.05$ \\
    Methane & - \\
    Agriculture & - \\
    \emph{Note} & Entries not listed are $0$. \\
    \midrule
    Costs (per agent) & Climate damage: $100$,\ Energy: $1\times 10^{-3}$,\ Methane: $10$,\ Agriculture: $10$,\ Adaptation: $10$ \\
    Prevention & Decay factor: $0.95$,\ Max prevention benefit: $0.0$ \\
    \bottomrule \\
  \end{tabular}
  \caption{Parameters used in MARL scenario (i).}
  \label{tab:marl_env_scenario_i}
\end{table*}

\begin{table*}[h]
  \centering
  \small
  \renewcommand{\arraystretch}{1.15}
  \setlength{\tabcolsep}{6pt}
  \begin{tabular}{p{3.2cm} p{12.2cm}}
    \toprule
    \textbf{Item} & \textbf{Specification} \\
    \midrule
    Agents ($N$) & $N=10$ agents with shares $S_i = [0.35, 0.15, 0.10, 0.05, 0.02, 0.01, 0.03, 0.14, 0.1, 0.05]$ \\
    Climate engines & Surrogates only\\
    Controlled gases & CO2\_FF, CO2\_AFOLU, CH4, N2O, SO2 \\
    \midrule
    \multicolumn{2}{l}{\textbf{Levers and levels}} \\
    Energy & Levels: $0.0,\ 0.5,\ 1.0$ \\
    Methane & Levels: $0.0,\ 0.5,\ 1.0$ \\
    Agriculture & Levels: $0.0,\ 0.5,\ 1.0$ \\
    Adaptation & Levels: $0.00,\ 0.03,\ 0.08$ \\
    \midrule
    \multicolumn{2}{l}{\textbf{Policy to emissions mapping (per-year $\Delta$ growth coefficients)}} \\
    Energy & CO2\_FF: $-0.05$, CH4: $-0.005$, N2O: $-0.005$, SO2: $-0.05$ \\
    Methane & CH4: $-0.04$ \\
    Agriculture & CO2\_AFOLU: $-0.04$, CH4: $-0.005$, N2O: $-0.03$ \\
    \emph{Note} & Entries not listed are $0$. \\
    \midrule
    Costs (per agent) & Climate damage: $[50,\ 50,\ 100,\ 100,\ 10,\ 25,\ 50,\ 1000,\ 1,\ 15]$ \\
                      & Energy: $[10^{-3},\ 10^{-2},\ 10^{-1},\ 10,\ 10^{-1},\ 10^{-3},\ 10^{-2},\ 10^{-1},\ 10,\ 10^{-1}]$ \\
                      & Methane: $[10^{-3},\ 10^{-2},\ 10,\ 10^{-1},\ 10^{-1},\ 2\times10^{-1},\ 5\times10^{-2},\ 10^{-1},\ 10,\ 10^{-1}]$ \\
                      & Agriculture: $[10^{-1},\ 10,\ 10^{-2},\ 10^{-3},\ 10^{-1},\ 10^{-3},\ 10,\ 100,\ 10,\ 10^{-1}]$ \\
                      & Adaptation: $[10,\ 10^{-1},\ 10^{-2},\ 10^{-3},\ 10^{-1},\ 10^{-3},\ 10^{-2},\ 10^{-1},\ 10,\ 10^{-1}]$ \\
    Prevention & Decay factor: $0.95$,\ Max prevention benefit: $0.5$ \\
    \bottomrule \\
  \end{tabular}
  \caption{Parameters used in MARL scenario (ii).}
  \label{tab:marl_env_scenario_ii}
\end{table*}

\clearpage

\noindent In addition to the implementation details, we provide additional details of the results of the MARL experiments.
In Table \ref{tab:marl_figure_overview} an overview of the additional figures is shown.

\begin{table*}[h]
  \centering
  \small
  \renewcommand{\arraystretch}{1.15}
  \setlength{\tabcolsep}{6pt}
  \begin{tabular}{l l p{0.55\textwidth}}
    \toprule
    \textbf{Topic} & \textbf{Scenario} & \textbf{Description of figure} \\
    \midrule
    Training time comparison & Homogeneous & \hyperref[fig:marl_training_time]{Fig.~\ref*{fig:marl_training_time} — Wall-clock training time for(surrogate vs.\ simulator)} \\
    Reward convergence & Homogeneous & \hyperref[fig:marl_reward_comparsion_homogeneous]{Fig.~\ref*{fig:marl_reward_comparsion_homogeneous} — Reward convergence per agent (surrogate vs.\ simulator)}\\
    Reward convergence & Heterogeneous & \hyperref[fig:marl_reward_heterogeneous]{Fig.~\ref*{fig:marl_reward_heterogeneous} — Reward convergence per agent (surrogate only)} \\
    Mean lever policies & Homogeneous & \hyperref[fig:lever_mean_consistency]{Fig.~\ref*{fig:lever_mean_consistency} — Mean lever convergence  (surrogate vs. simulator)} \\
    Mean lever policies & Heterogeneous & \hyperref[fig:mean_lever_consistency_heterogenous]{Fig.~\ref*{fig:mean_lever_consistency_heterogenous} — Mean lever efforts convergence (surrogates only)} \\
    Per-agent levers (GRU) & Homogeneous & \hyperref[fig:per_agent_levers_gru]{Fig.~\ref*{fig:per_agent_levers_gru} — Per-agent lever convergence (SCM vs.\ GRU)} \\
    Per-agent levers (LSTM) & Homogeneous & \hyperref[fig:per_agent_levers_lstm]{Fig.~\ref*{fig:per_agent_levers_lstm} — Per-agent lever convergence (SCM vs.\ LSTM)} \\
    Per-agent levers (TCN) & Homogeneous & \hyperref[fig:per_agent_levers_tcn]{Fig.~\ref*{fig:per_agent_levers_tcn} — Per-agent lever convergence (SCM vs.\ TCN)} \\
    Per-agent levers (GRU) & Heterogeneous  & \hyperref[fig:per_agent_levers_heterogeneous_gru]{Fig.~\ref*{fig:per_agent_levers_heterogeneous_gru} — Per-agent lever convergence (heterogeneous, GRU)} \\
    Per-agent levers (LSTM) & Heterogeneous & \hyperref[fig:per_agent_levers_heterogeneous_lstm]{Fig.~\ref*{fig:per_agent_levers_heterogeneous_lstm} — Per-agent lever convergence (heterogeneous, LSTM)} \\
    Per-agent levers (TCN) & Heterogeneous & \hyperref[fig:per_agent_levers_heterogeneous_tcn]{Fig.~\ref*{fig:per_agent_levers_heterogeneous_tcn} — Per-agent lever convergence (heterogeneous, TCN)} \\
    \bottomrule \\
  \end{tabular}
  \caption{Overview of additional figures for MARL experiments.}
  \label{tab:marl_figure_overview}
\end{table*}

\begin{figure*}[ht]
  \centering
  
  \begin{subfigure}{0.5\textwidth}
    \centering
    \includegraphics[width=0.95\linewidth]{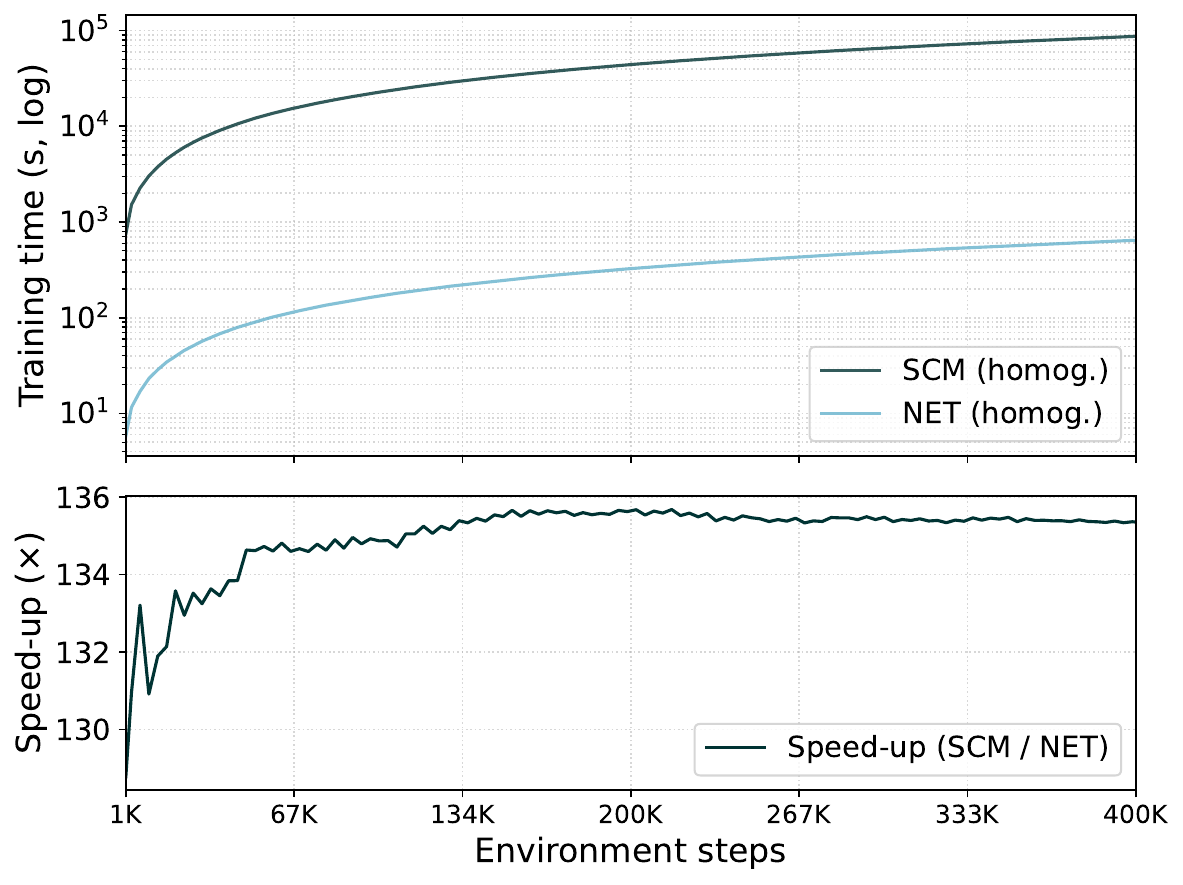}
    \caption{GRU}
  \end{subfigure}
  
  \vspace{0.5em}

  \begin{subfigure}{0.5\textwidth}
    \centering
    \includegraphics[width=0.95\linewidth]{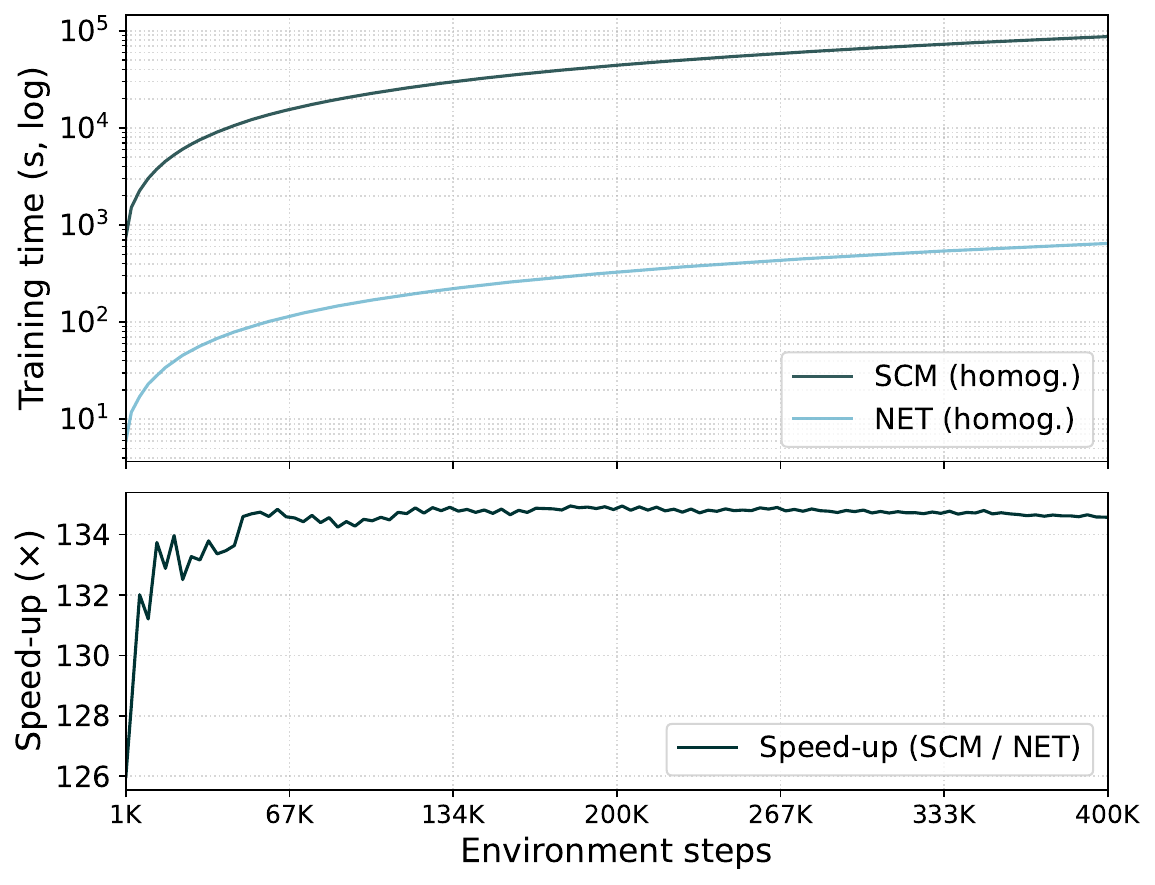}
    \caption{LSTM}
  \end{subfigure}

  \vspace{0.5em}
  
  \begin{subfigure}{0.5\textwidth}
    \centering
    \includegraphics[width=0.95\linewidth]{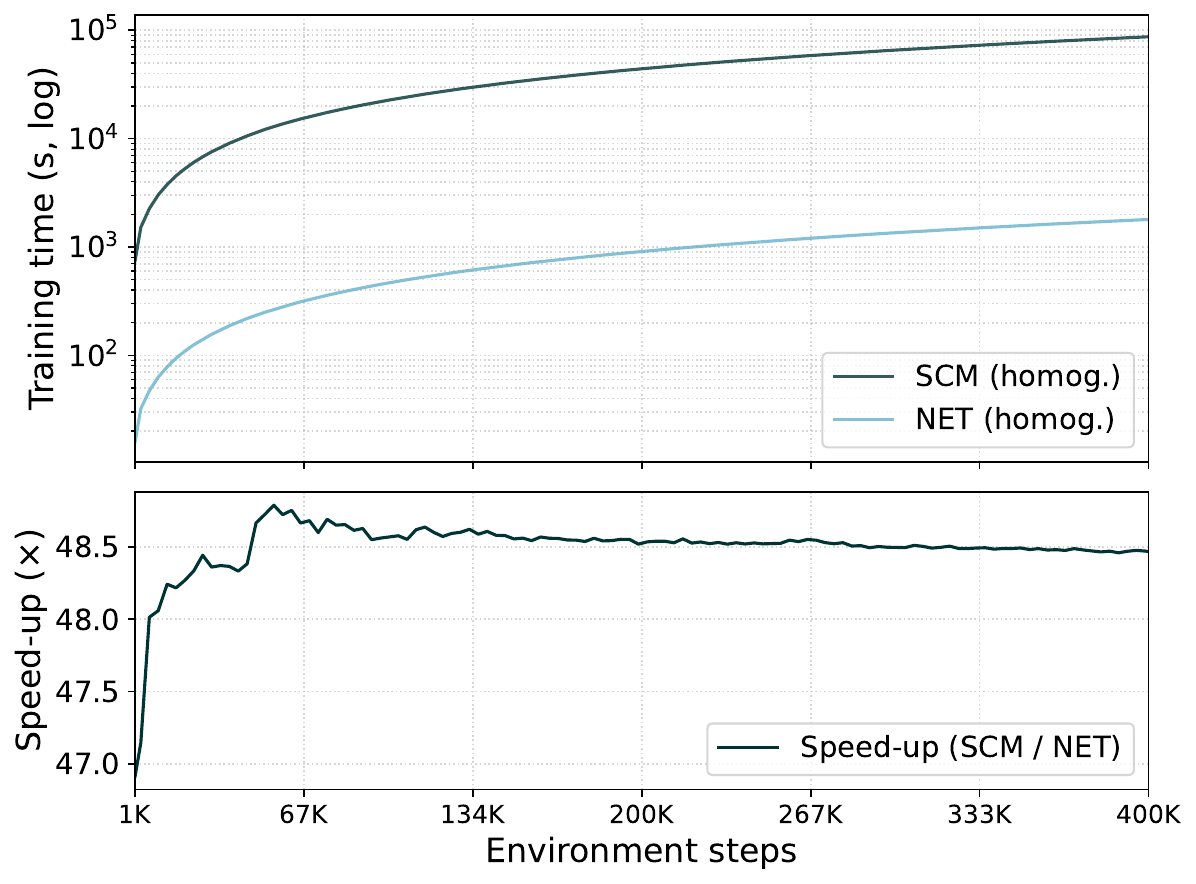}
    \caption{TCN}
  \end{subfigure}
  
  \caption{Comparison of training time during MARL for the three surrogate architectures (GRU, LSTM, and TCN) relative to CICERO-SCM in the homegeneous scenario (i). 
  Each panel shows the wall-clock training time (log scale) and the corresponding speed-up achieved by replacing the simulator with the surrogate model.}
  \label{fig:marl_training_time}
  
\end{figure*}

\begin{figure*}[ht]
  \centering
  
  \begin{subfigure}{0.65\textwidth}
    \centering
    \includegraphics[width=0.95\linewidth]{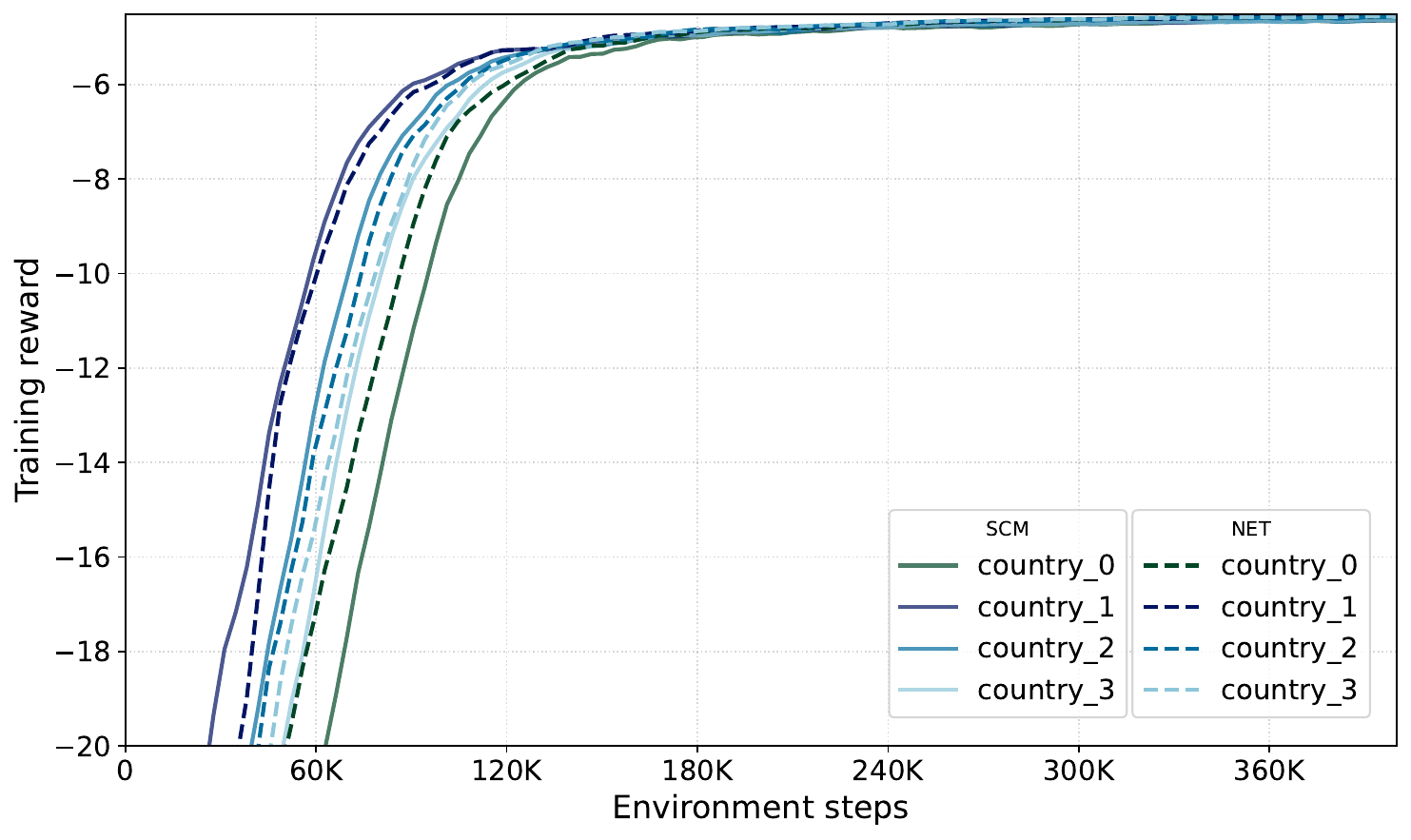}
    \caption{GRU}
  \end{subfigure}

  \vspace{0.5em}
  
  \begin{subfigure}{0.65\textwidth}
    \centering
    \includegraphics[width=0.95\linewidth]{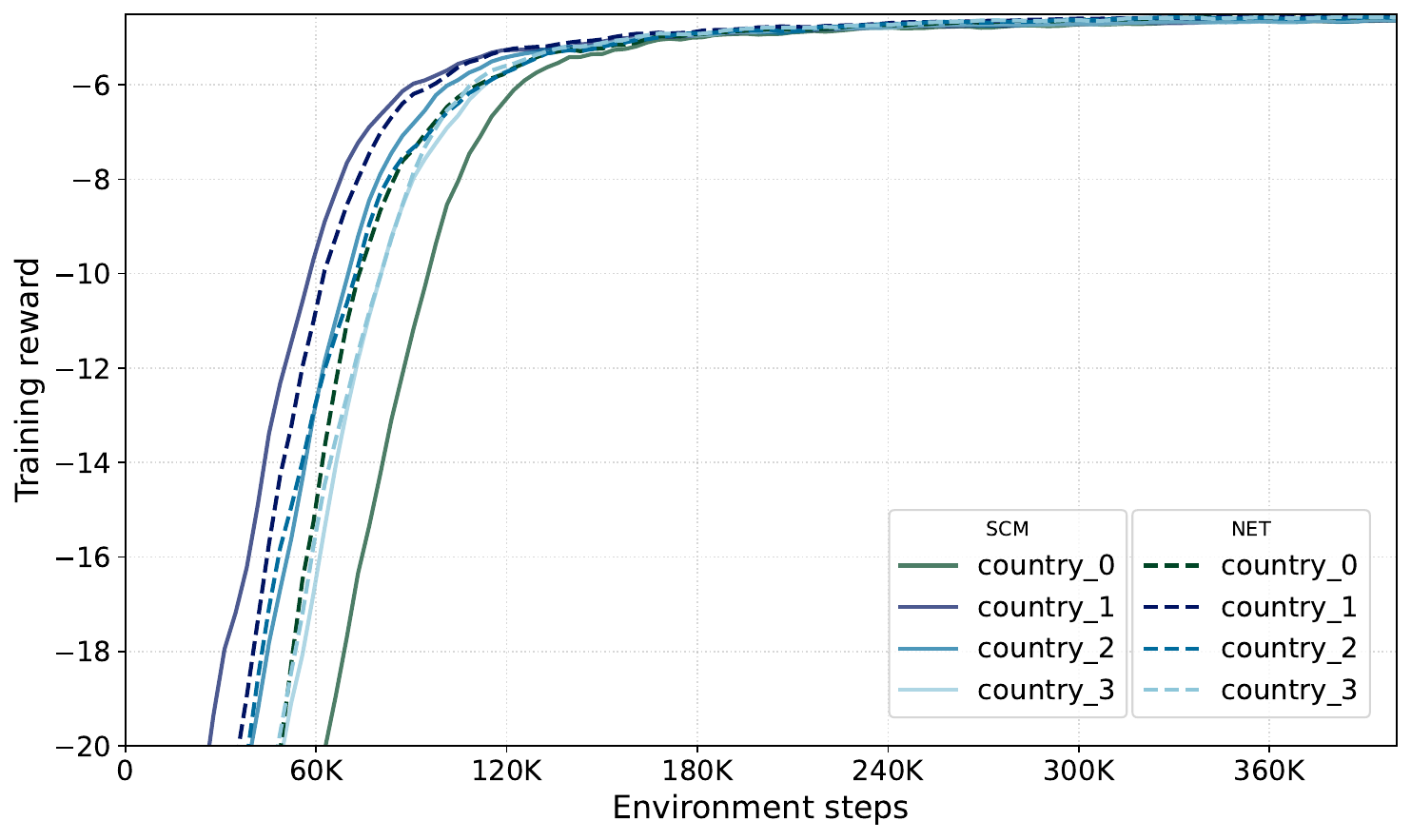}
    \caption{LSTM}
  \end{subfigure}

  \vspace{0.5em}
  
  \begin{subfigure}{0.65\textwidth}
    \centering
    \includegraphics[width=0.95\linewidth]{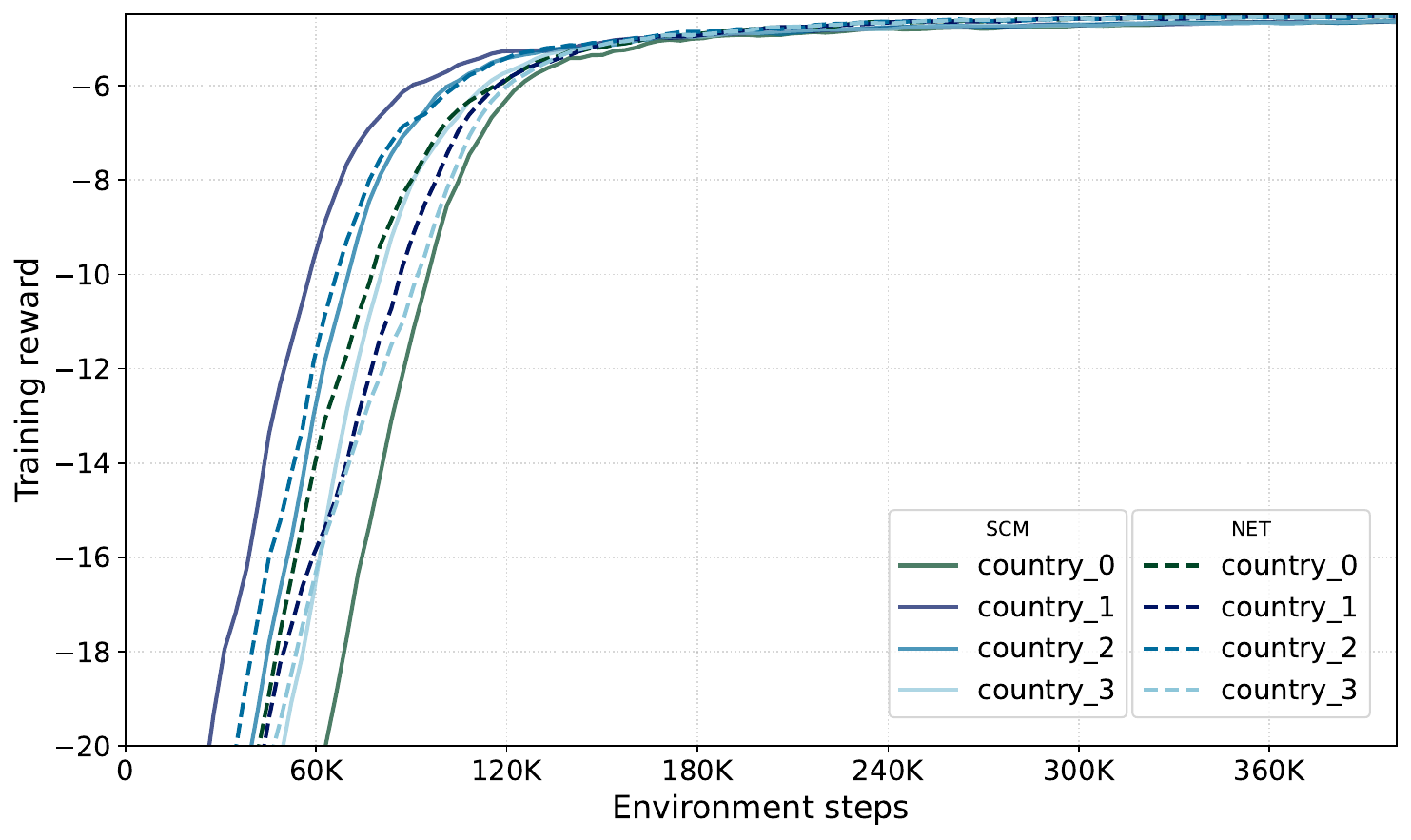}
    \caption{TCN}
  \end{subfigure}
  
    \caption{Comparison of training reward per agent for the three surrogate architectures (GRU, LSTM, and TCN) relative to CICERO-SCM in the homogeneous scenario (i).
    Each panel shows the evolution of agents’ rewards across environment steps, comparing trajectories obtained with the surrogate (dashed) and the simulator (solid).}  \label{fig:marl_reward_comparsion_homogeneous}
\end{figure*}

\begin{figure*}[ht]
  \centering
  
  \begin{subfigure}{0.65\textwidth}
    \centering
    \includegraphics[width=0.95\linewidth]{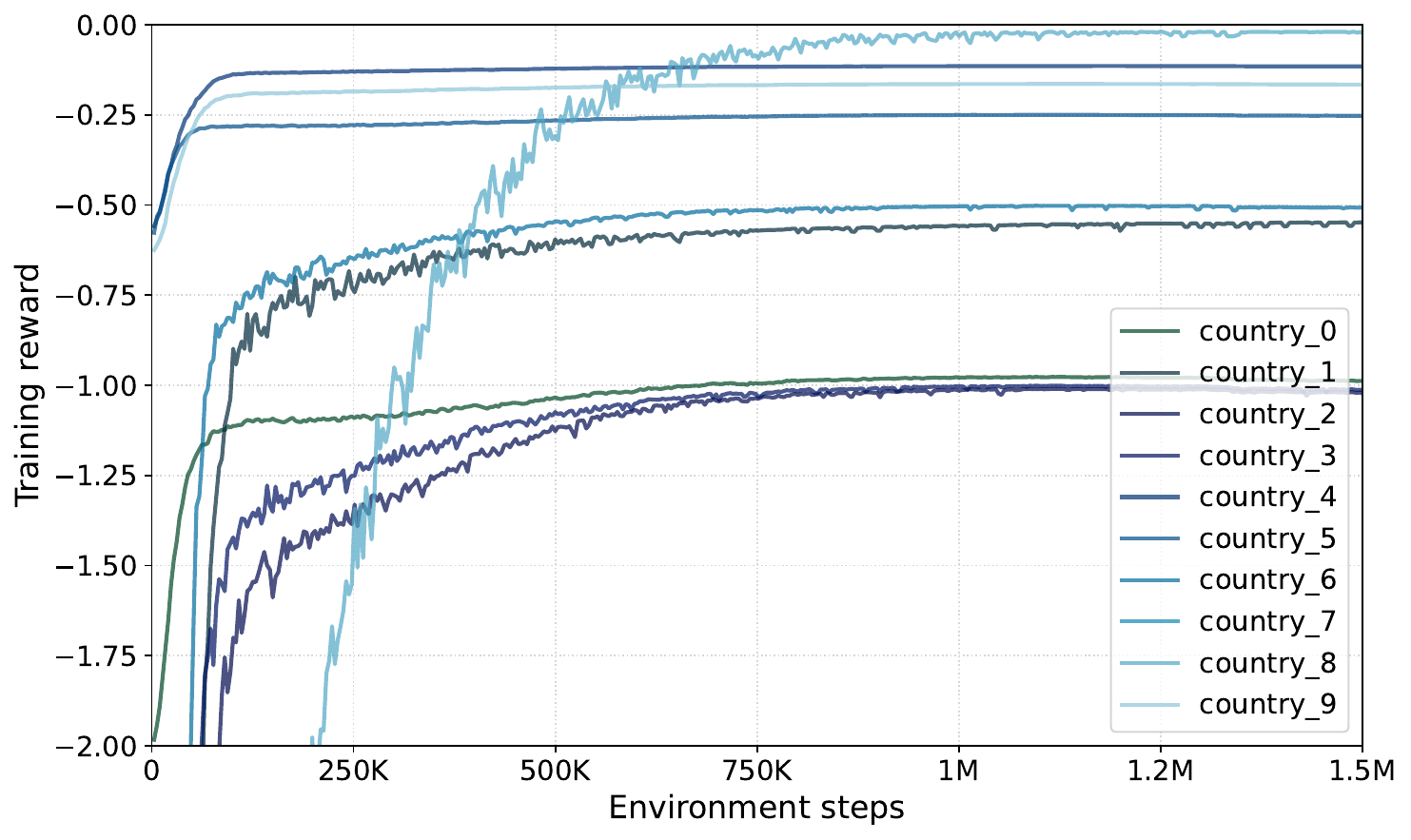}
    \caption{GRU}
  \end{subfigure}
  
  \vspace{0.5em}

  \begin{subfigure}{0.65\textwidth}
    \centering
    \includegraphics[width=0.95\linewidth]{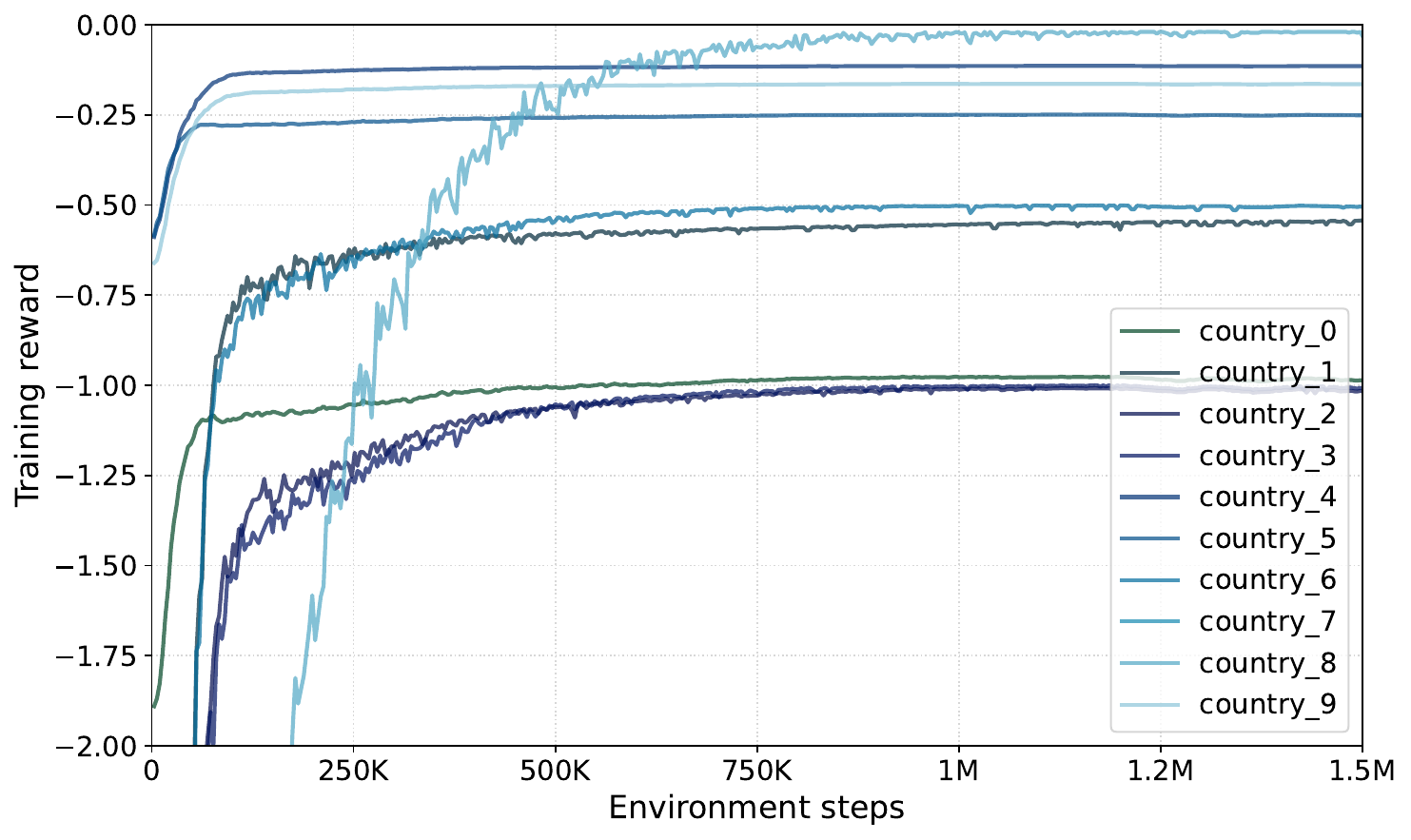}
    \caption{LSTM}
  \end{subfigure}

  \vspace{0.5em}
  
  \begin{subfigure}{0.65\textwidth}
    \centering
    \includegraphics[width=0.95\linewidth]{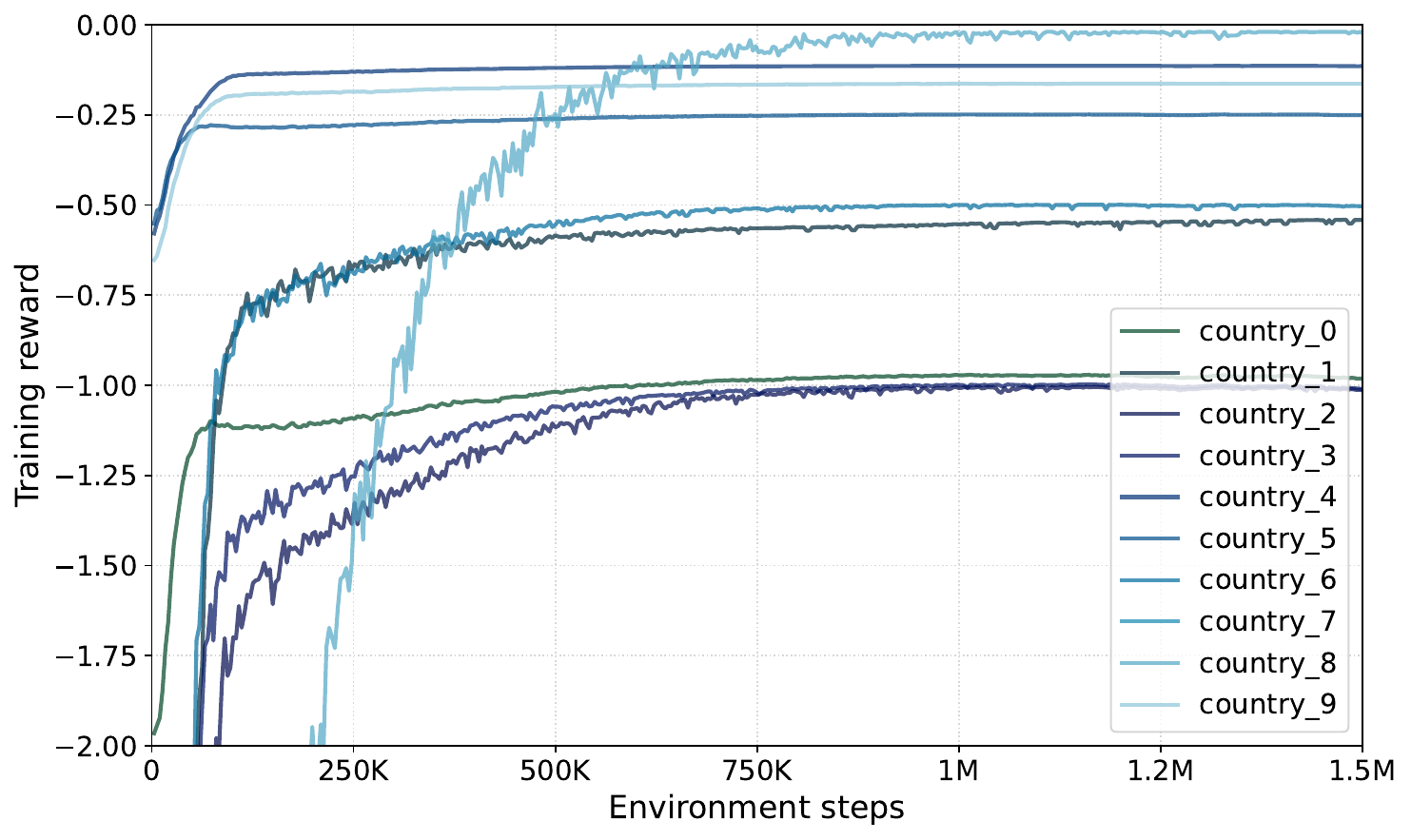}
    \caption{TCN}
  \end{subfigure}
  
  \caption{Training reward per agent in the heterogeneous scenario (ii) for GRU, LSTM, and TCN surrogates. 
  Each panel shows reward trajectories over environment steps.}  \label{fig:marl_reward_heterogeneous}
\end{figure*}

\begin{figure*}[ht]
  \centering
  \begin{subfigure}{0.65\textwidth}
    \centering
    \includegraphics[width=0.95\linewidth]{plots/lever_mean_consistency_gru_vs_cicero.pdf}
    \caption{GRU}
  \end{subfigure}
  
  \vspace{0.5em}

  \begin{subfigure}{0.65\textwidth}
    \centering
    \includegraphics[width=0.95\linewidth]{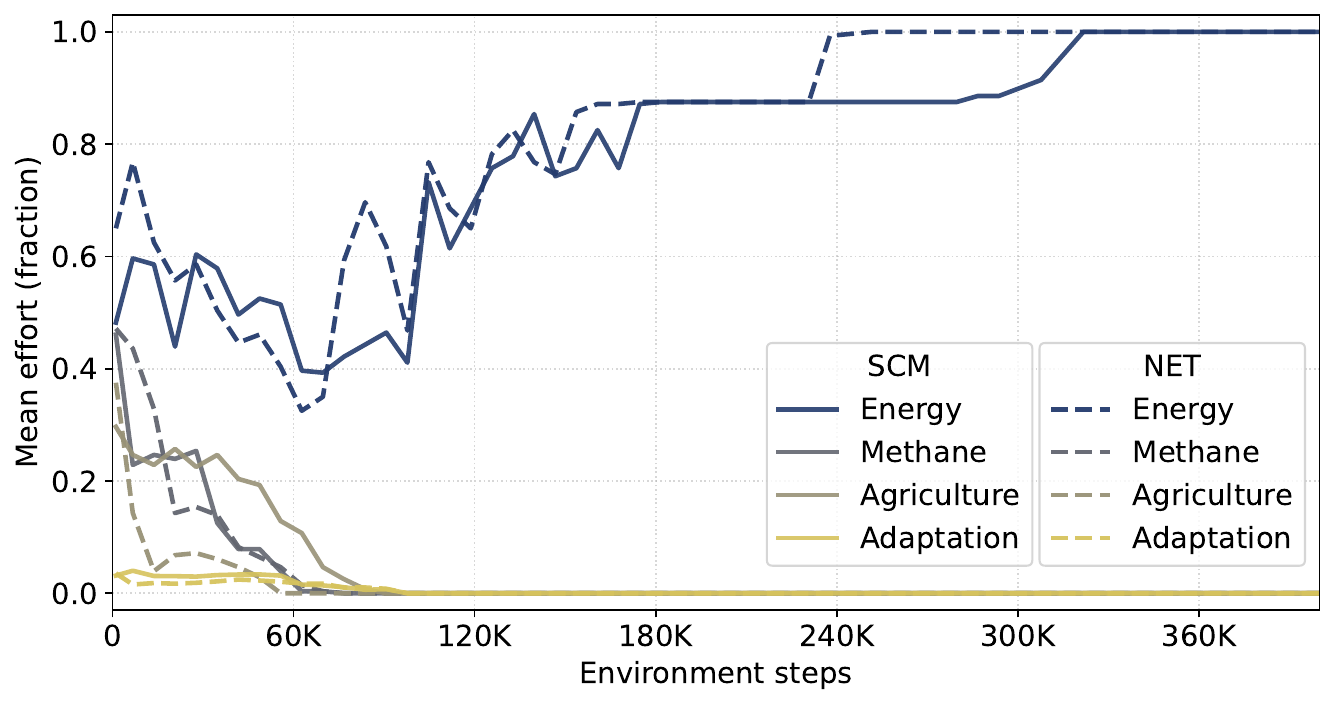}
    \caption{LSTM}
  \end{subfigure}

  \vspace{0.5em}
  \begin{subfigure}{0.65\textwidth}
    \centering
    \includegraphics[width=0.95\linewidth]{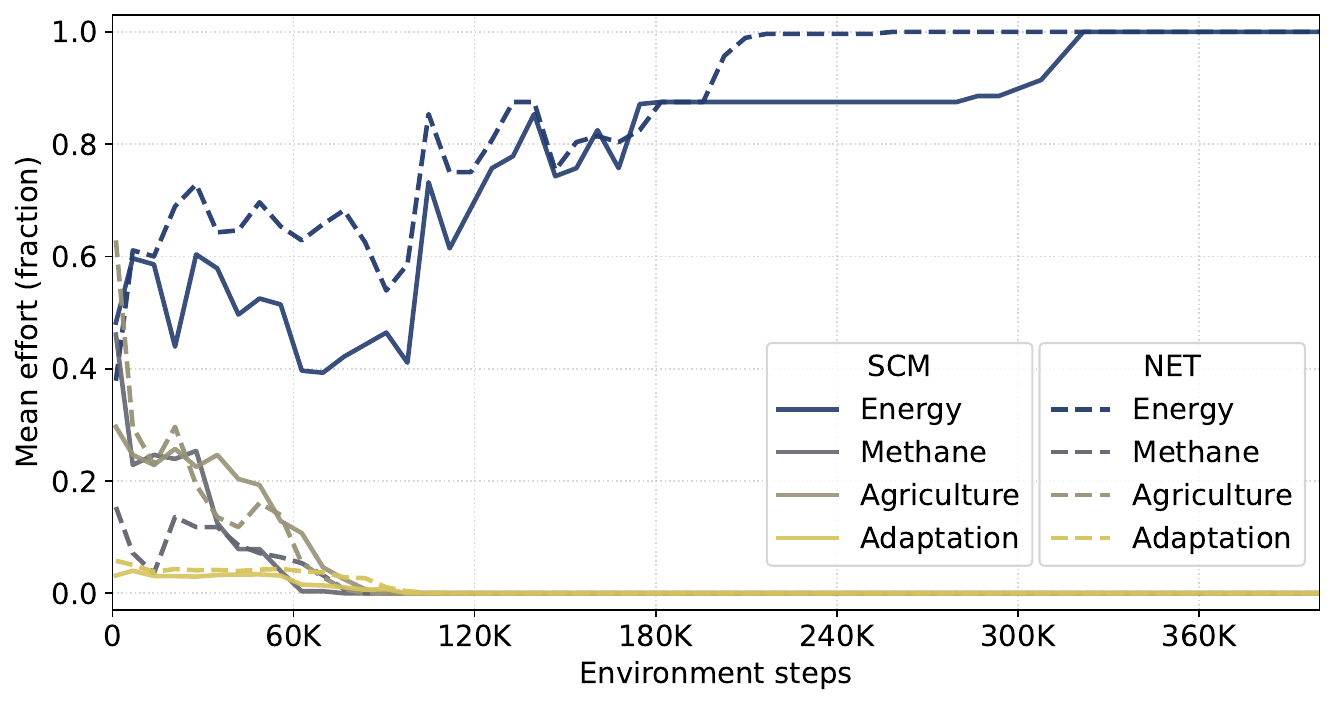}
    \caption{TCN}
  \end{subfigure}
  \caption{Comparison of per-lever mean policy trajectories in the homogeneous scenario (i) between CICERO-SCM and the surrogate models (GRU, LSTM, and TCN). 
  Each line shows the evolution of average lever efforts across agents across episodes for both engines, with solid lines representing CICERO-SCM and dashed lines the corresponding surrogate. 
  The close alignment indicates that the surrogates reproduce the learned policy dynamics of the simulator.}
  \label{fig:lever_mean_consistency}
\end{figure*}

\begin{figure*}[ht]
  \centering
  
  \begin{subfigure}{0.65\textwidth}
    \centering
    \includegraphics[width=0.95\linewidth]{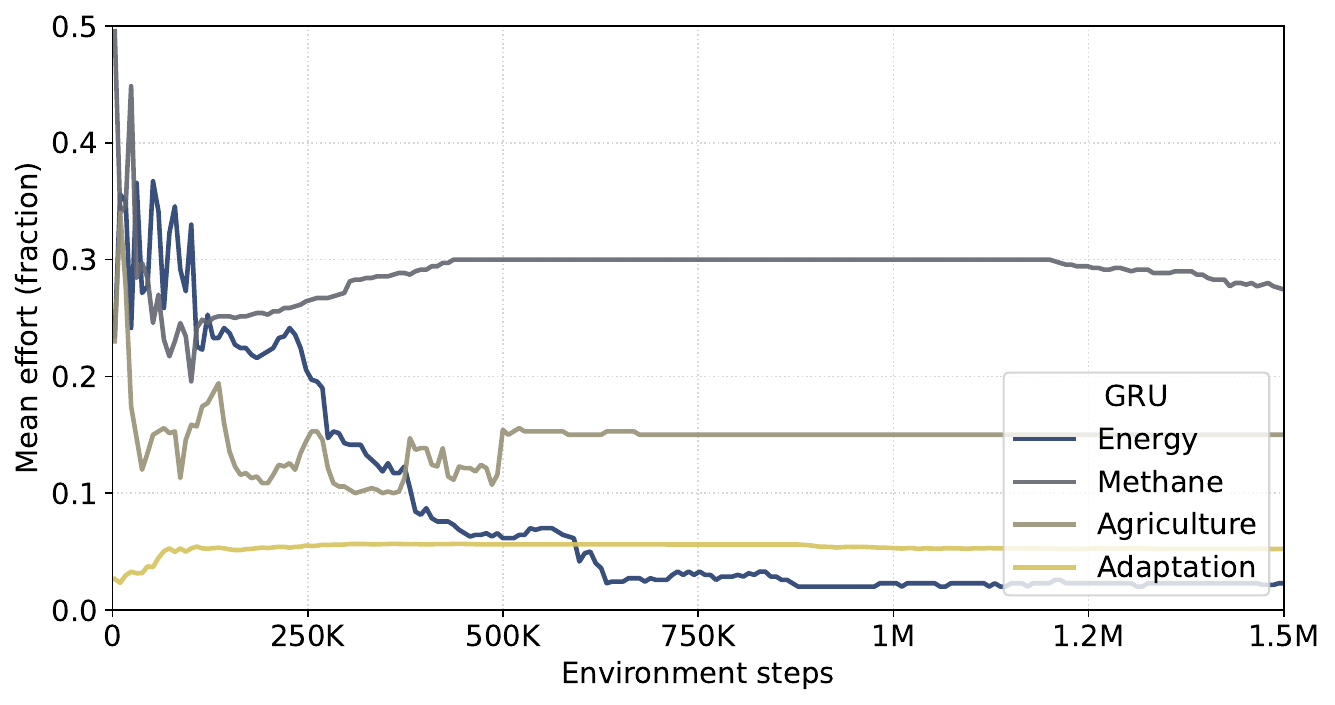}
    \caption{GRU}
  \end{subfigure}
  
  \vspace{0.5em}
  
  \begin{subfigure}{0.65\textwidth}
    \centering
    \includegraphics[width=0.95\linewidth]{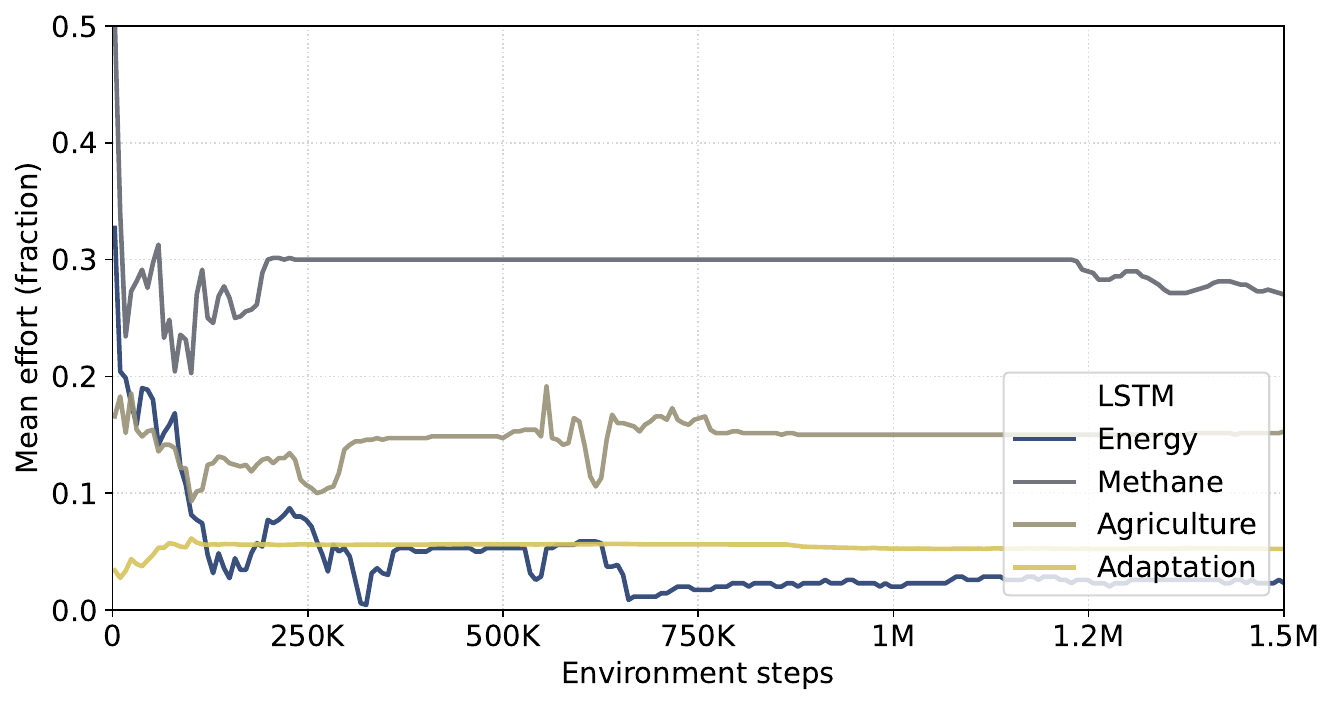}
    \caption{LSTM}
  \end{subfigure}

  \vspace{0.5em}
  
  \begin{subfigure}{0.65\textwidth}
    \centering
    \includegraphics[width=0.95\linewidth]{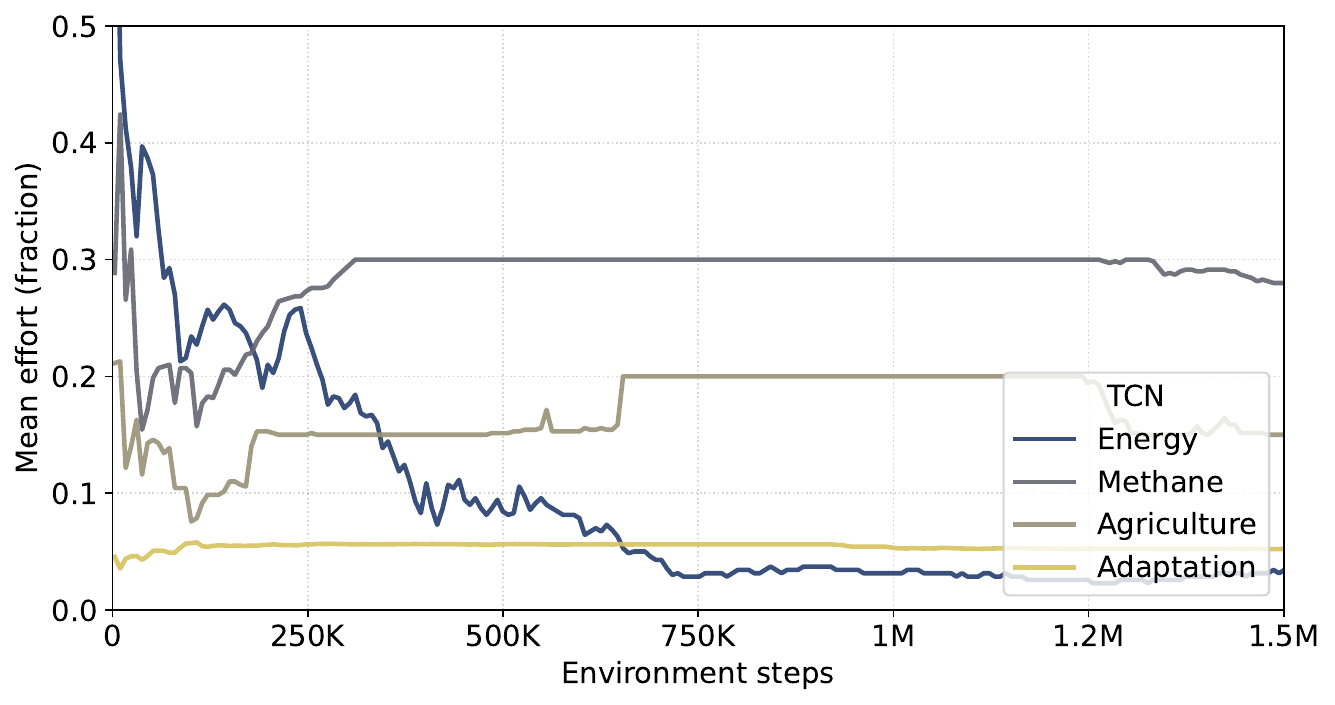}
    \caption{TCN}
  \end{subfigure}
  
  \caption{Per-lever mean policy trajectories in the heterogeneous scenario (ii) for the GRU, LSTM, and TCN surrogates. 
  Each line shows the evolution of average lever efforts across agents across episodes.}
  
  \label{fig:mean_lever_consistency_heterogenous}
\end{figure*}

\begin{figure*}[p]
  \centering
  \includegraphics[width=0.95\linewidth]{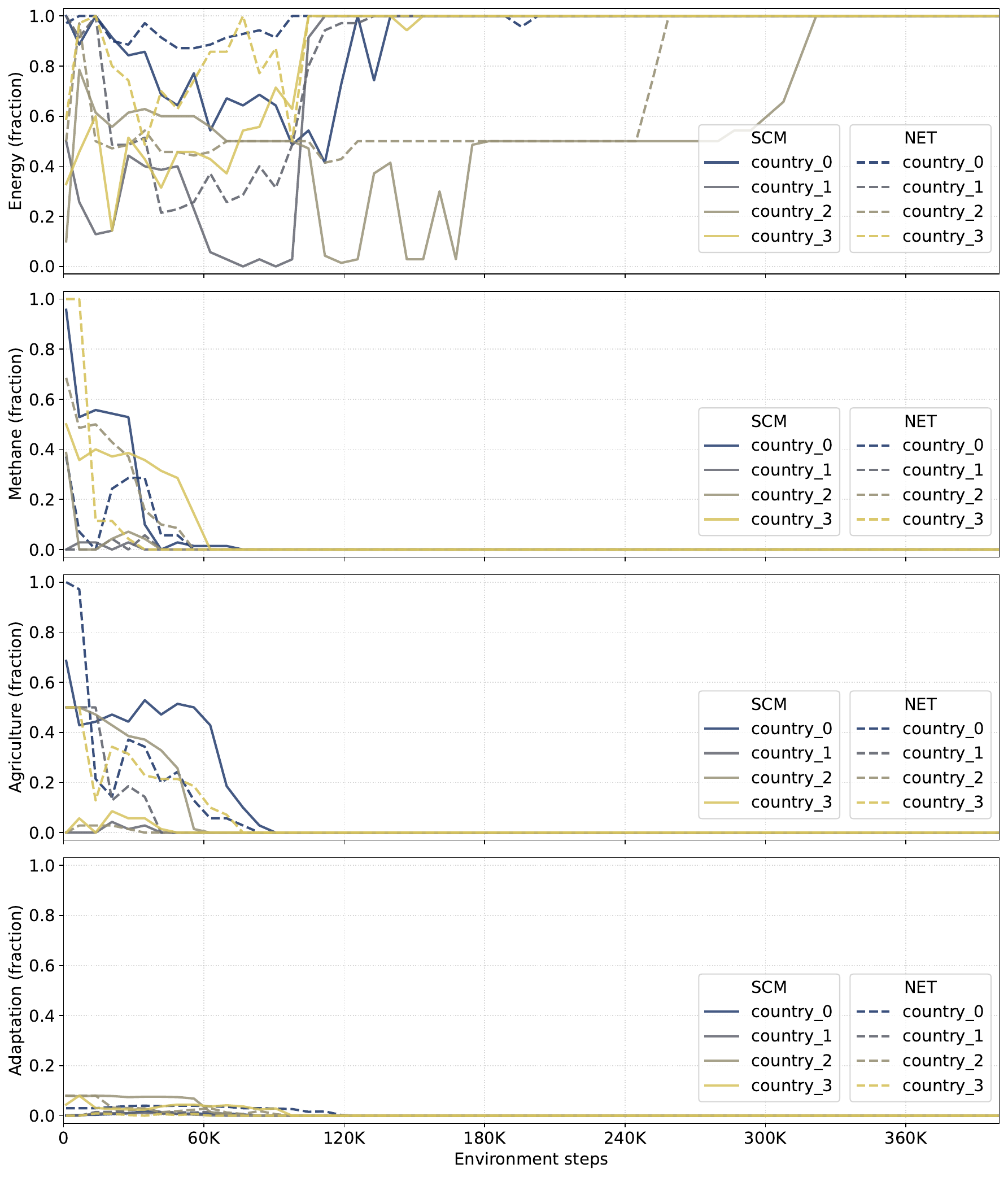}
  \caption{Trajectories of per-agent mean lever effect across episodes shown across environment steps for the homogeneous scenario (i) under CICERO-SCM and GRU surrogate.}
  \label{fig:per_agent_levers_gru}
\end{figure*}

\begin{figure*}[p]
  \centering
  \includegraphics[width=0.95\linewidth]{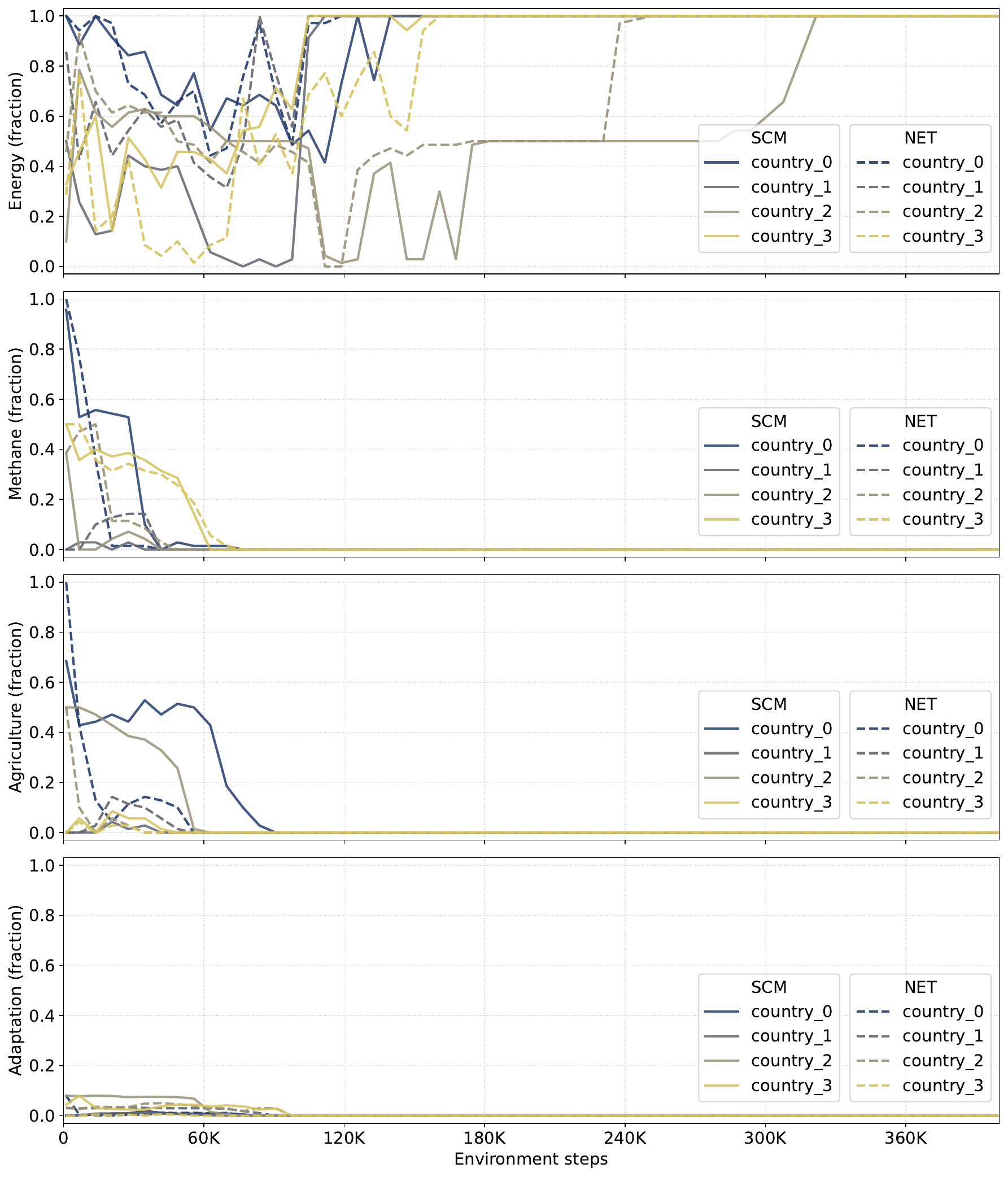}
  \caption{Trajectories of per-agent mean lever effect across episodes shown across environment steps for the homogeneous scenario (i) under CICERO-SCM and LSTM surrogate.}
  \label{fig:per_agent_levers_lstm}
\end{figure*}

\begin{figure*}[p]
  \centering
  \includegraphics[width=0.95\linewidth]{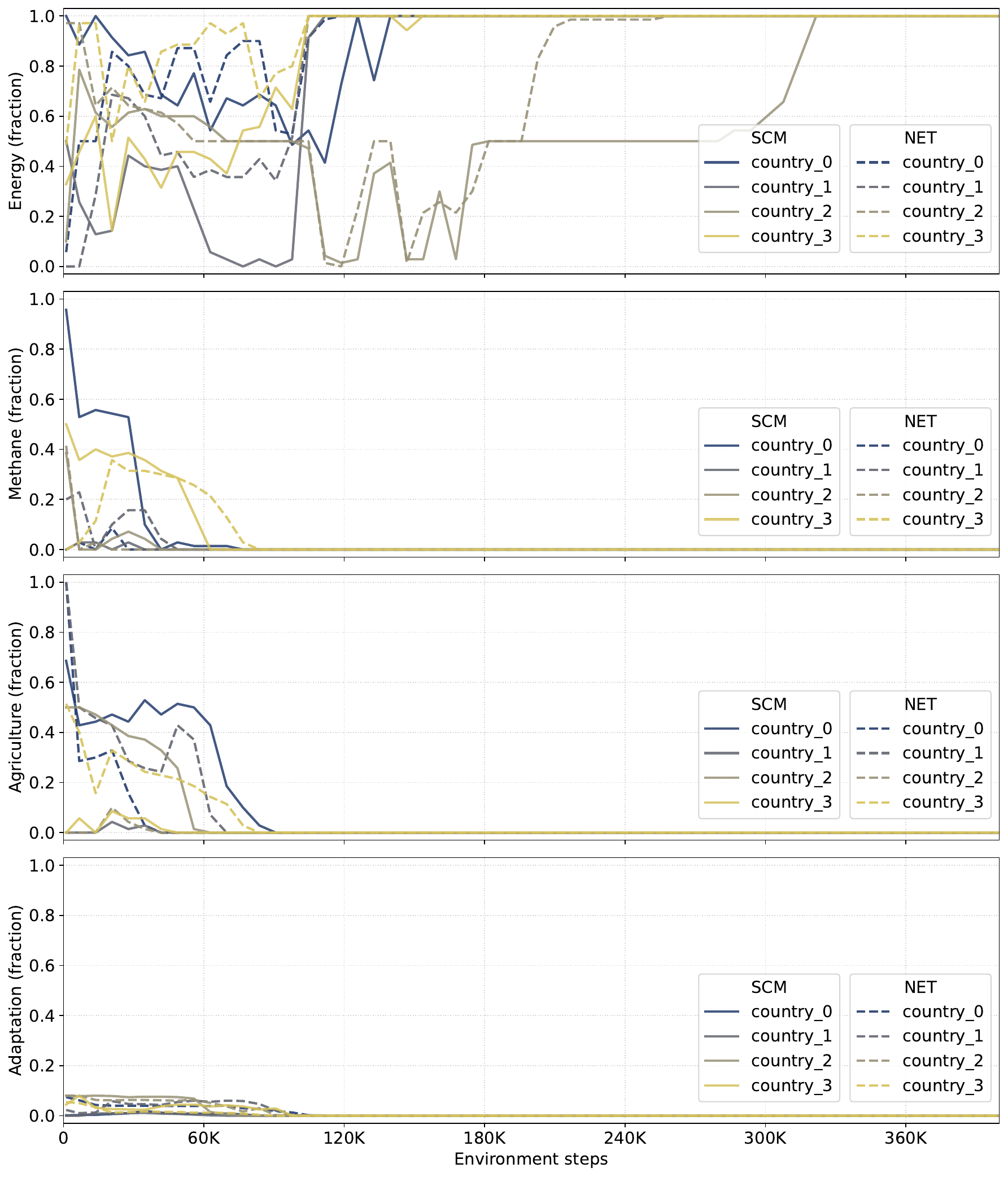}
  \caption{Trajectories of per-agent mean lever effect across episodes shown across environment steps for the homogeneous scenario (i) under CICERO-SCM and TCN surrogate.}
  \label{fig:per_agent_levers_tcn}
\end{figure*}

\begin{figure*}[p]
  \centering
  \includegraphics[width=0.95\linewidth]{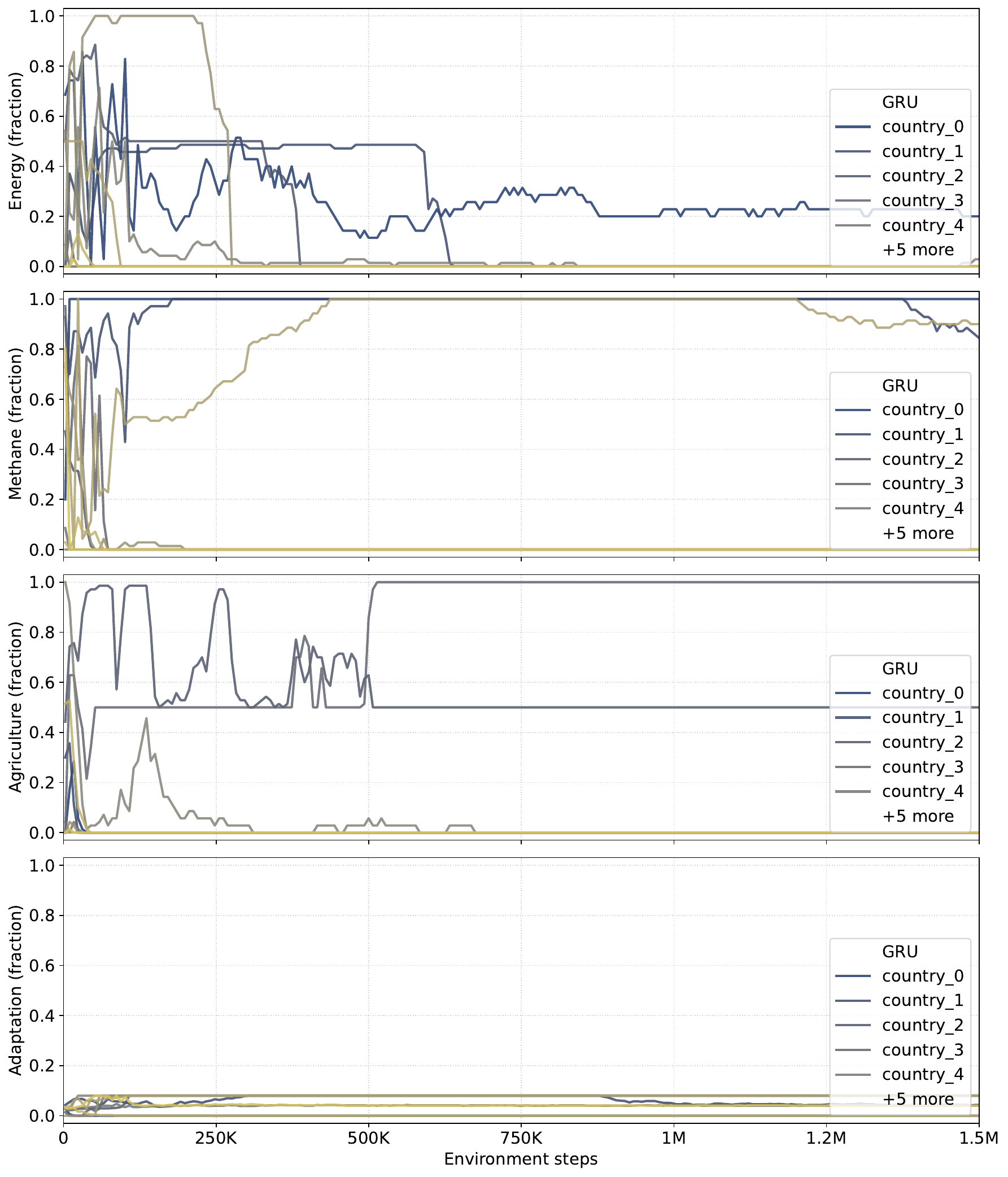}
  \caption{Trajectories of per-agent mean lever effect for heterogeneous scenario (ii) across episodes shown across environment steps under CICERO-SCM and GRU surrogate.}
  \label{fig:per_agent_levers_heterogeneous_gru}
\end{figure*}

\begin{figure*}[p]
  \centering
  \includegraphics[width=0.95\linewidth]{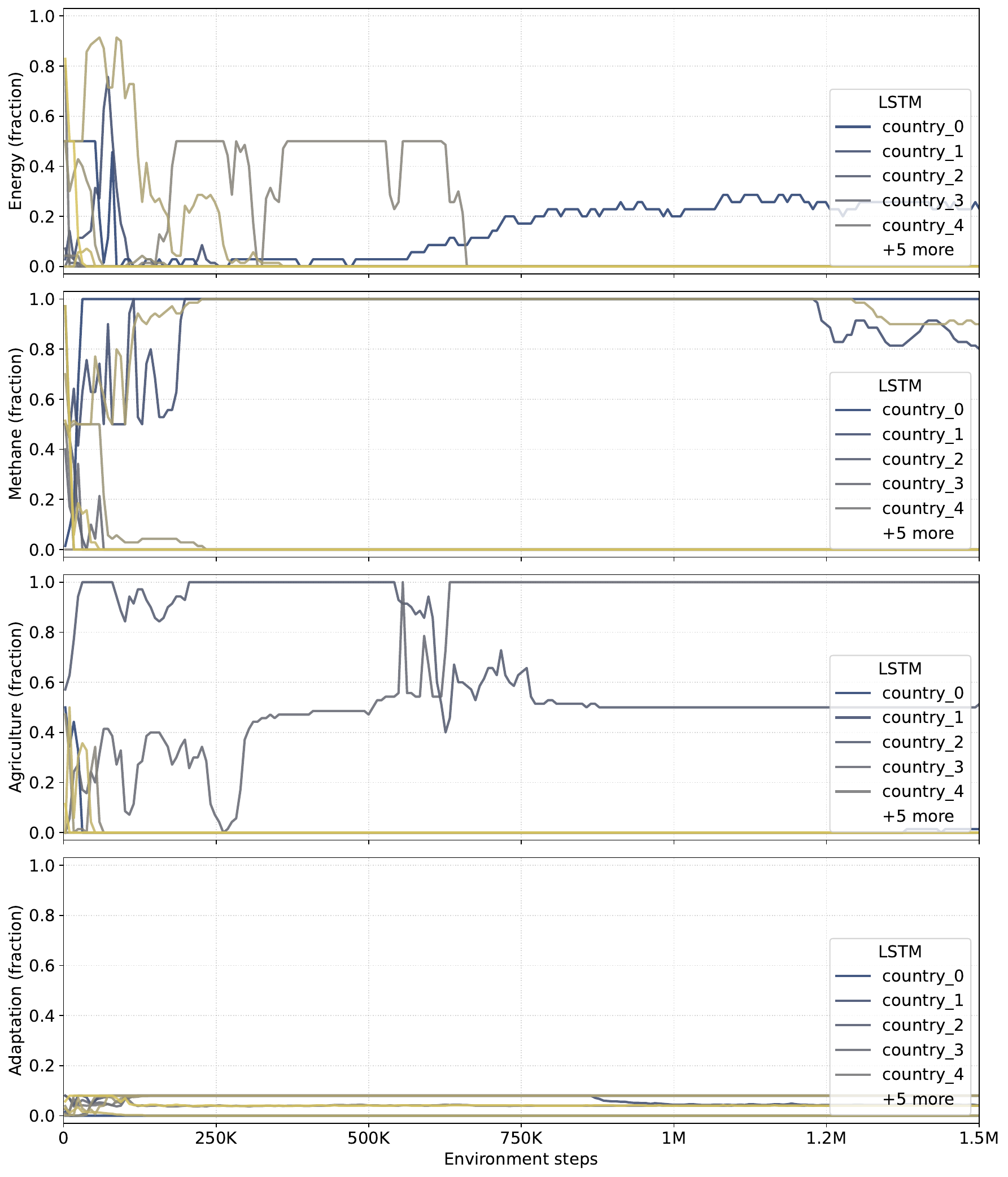}
  \caption{Trajectories of per-agent mean lever effect for heterogeneous scenario (ii) across episodes shown across environment steps under CICERO-SCM and LSTM surrogate.}
  \label{fig:per_agent_levers_heterogeneous_lstm}
\end{figure*}

\begin{figure*}[p]
  \centering
  \includegraphics[width=0.95\linewidth]{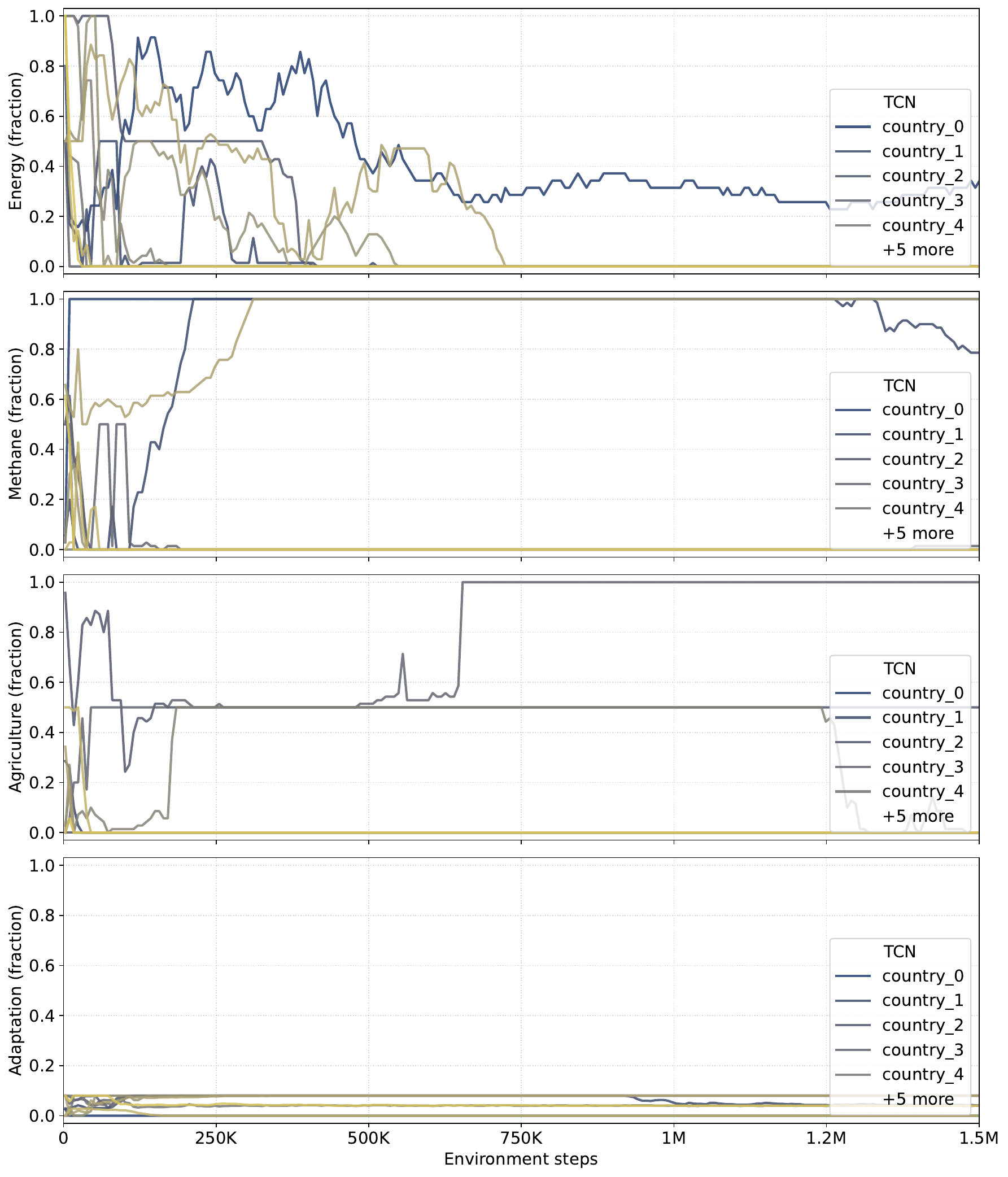}
  \caption{Trajectories of per-agent mean lever effect for heterogeneous scenario (ii) across episodes shown across environment steps under CICERO-SCM and TCN surrogate.}
  \label{fig:per_agent_levers_heterogeneous_tcn}
\end{figure*}

\end{document}